\documentclass[runningheads]{llncs}

\usepackage[T1]{fontenc}
\usepackage{graphicx}
\usepackage{subcaption}     % Para que funcione el subfigure
\usepackage[dvipsnames]{xcolor} % Este para definecolor, pero no hace la tabla
\usepackage{tikz}               % Este se usa en la tabla: la herramienta más chida para crear elementos gráficos directamente O.O
\usepackage{multirow}           % Este para hacer eso :v como combinar celdas en Excel
\usepackage{booktabs}           % Este es el que pone las rayitas

\usepackage{float}
\usepackage{tabu}
\usepackage{multirow}

\definecolor{mycolor0}{RGB}{0, 0, 0}

\definecolor{mycolor1}{RGB}{130, 76, 0}
\definecolor{mycolor2}{RGB}{0, 102, 0}
\definecolor{mycolor3}{RGB}{112, 103, 87}
\definecolor{mycolor4}{RGB}{112, 150, 146}
\definecolor{mycolor5}{RGB}{128, 64, 128}
\definecolor{mycolor6}{RGB}{107, 142, 35}
\definecolor{mycolor7}{RGB}{48, 41, 30}
\definecolor{mycolor8}{RGB}{0, 50, 89}
\definecolor{mycolor9}{RGB}{70, 70, 70}
\definecolor{mycolor10}{RGB}{190, 153, 153}
\definecolor{mycolor11}{RGB}{153, 153, 153}
\definecolor{mycolor12}{RGB}{28, 42, 168}
\definecolor{mycolor13}{RGB}{102, 102, 156}
\definecolor{mycolor14}{RGB}{254, 228, 12}
\definecolor{mycolor15}{RGB}{254, 148, 12}
\definecolor{mycolor16}{RGB}{119, 11, 32}
\definecolor{mycolor17}{RGB}{51, 51, 0}
\definecolor{mycolor18}{RGB}{190, 150, 190}
\definecolor{mycolor19}{RGB}{2, 135, 115}
\definecolor{mycolor20}{RGB}{102, 51, 0}
\definecolor{mycolor21}{RGB}{9, 143, 150}
\definecolor{mycolor22}{RGB}{255, 0, 0}
\definecolor{mycolor23}{RGB}{255, 22, 96}
\definecolor{mycolorR0}{RGB}{0, 0, 255}
\definecolor{mycolorR1}{RGB}{0, 255, 255}
\definecolor{mycolorR2}{RGB}{0, 255, 0}
\definecolor{mycolorR3}{RGB}{255, 255, 0}
\definecolor{mycolorR4}{RGB}{255, 128, 0}
\definecolor{mycolorR5}{RGB}{255, 0, 0}

\begin{document}

\title{Risk Assessment for Autonomous Landing in Urban Environments using Semantic Segmentation}
% If the paper title is too long for the running head, you can set an abbreviated paper title here
\titlerunning{Risk Assessment for Autonomous Landing}

\author{Jesús Alejandro Loera Ponce\inst{1}\orcidID{0000-1111-2222-3333} \and
Diego A. Mercado-Ravell*\inst{1,2}\orcidID{1111-2222-3333-4444} \and
Israel Becerra-Durán\inst{2}\orcidID{2222--3333-4444-5555} \and
Luis Manuel Valentin-Coronado\inst{3}\orcidID{3333-4444-5555-6666}}

\authorrunning{Loera Ponce et al.}
% First names are abbreviated in the running head.
% If there are more than two authors, 'et al.' is used.

\institute{Center for Research in Mathematics, CIMAT AC, Zacatecas ZAC 98160, MEXICO
\and
Investigadores CONAHCYT, Centro de Investigación en Matemáticas, A.C., %Zacatecas ZAC 98160, MEXICO
Guanajuato GTO 36023, MEXICO
%\email{diego.mercado@cimat.mx}\\
%\url{http://www.springer.com/gp/computer-science/lncs}
\and
Investigadores CONAHCYT, Centro de Investigaciones en Óptica, A.C., Aguascalientes AGS 20200, MEXICO}

\maketitle              % typeset the header of the contribution

\begin{abstract}
In this paper, we address the vision-based autonomous landing problem in complex urban environments using deep neural networks for semantic segmentation and risk assessment. %We propose the use of a state-of-the-art visual transformer network, the SegFormer, for semantic segmentation of complex unstructured urban environments, which in turn provides valuable information that can be used for smart autonomous landing missions, particularly during emergency landing situations due to system failures or human errors. 
We propose employing the SegFormer, a state-of-the-art visual transformer network, for the semantic segmentation of complex, unstructured urban environments. This approach yields valuable information that can be utilized in smart autonomous landing missions, particularly in emergency landing scenarios resulting from system failures or human errors. %Aerial views taken from an Unmanned Aerial Vehicle (UAV) are segmented into %$23$ different classes of interest, including people, animals, cars, water, grass, pavement, vegetation, etc.
The assessment is done in real-time flight, when images of an RGB camera at the Unmanned Aerial Vehicle (UAV) are segmented with the SegFormer into 
the most common classes found in urban environments. These classes are then mapped into a level of risk, considering in general, potential material damage, damaging the drone itself %, and, in general, hurting people. 
and endanger people. The proposed strategy is validated through several %cases of study
case studies, demonstrating the huge potential of semantic segmentation-based strategies %as a powerful sensor for 
to determining the safest landing areas for autonomous emergency landing, which we believe will help unleash the full potential of %drones in civil missions in smart cities.
UAVs on civil applications within urban areas.

\end{abstract}

\section{Introduction}
The rapid development of Unmanned Aerial Vehicles (UAVs) has allowed an increase in their use in all sorts of applications, especially civilian, ranging from simple entertainment through surveillance and infrastructure inspection, crossing the areas of professional photography, precision agriculture, and aerial transportation, among others~\cite{Abualigah2021,Teixeira2023}. However, even with the significant progress on UAVs, their deployment in populated areas has been hindered due to
%security
\textcolor{black}{safety} concerns. %Most applications are restricted to areas devoid of people due to the latent risk that a flying device can pose on people, not to mention material damage, in the event of malfunction or human errors, limiting the full potential of UAVs within cities \cite{FAA2021}.
\textcolor{black}{Most applications are restricted to areas devoid of people due to the latent risk that flying objects pose on them or on property in the event of malfunction or human errors, limiting the full potential of UAVs within urban spaces.}
% such as the Specific Operations Risk Assessment (SORA) \cite{}

\subsection{Unmanned Aerial Vehicle Regulations}

\textcolor{black}{Acordingly, regulations require to present risk mitigation measures to local authorities to allow flight missions in civilian spaces. %However, for the most part, strategies require to not fly around people and/or to have an emergency response plans.
In most cases, strategies entail avoiding densely populated areas and/or implementing emergency response plans.}
%, but currently most strategies is to avoid  On the Unmanned Aircraft Systems to undergo several examinations and protocols before authorizing the use of

\textcolor{black}{%For the European Union Aviation Safety Agency (EASA) requires for UAVs in the "Open" and "Specific"
The European Union Aviation Safety Agency (EASA)~\cite{EASA2024} adopt a risk-based approach that do not distinguish between leisure or commercial activities, but the weight, specifications and operation intended to undertake. To fly over people, the EASA only authorizes drones weighting under $250g$, but never under "assemblies of people". In any case, it suggest to minimize the time flying over people. For CE class 2, the general rule is to keep the UAV at a lateral distance from any uninvolved person that is not less than the height at which the drone is flying and never fly closer than 30 meters horizontally from any uninvolved person. For the rest of categories, it is needed to ensure that no uninvolved people are present within range of the operation.}

\textcolor{black}{For the U.S. Federal Aviation Administration (FAA)~\cite{FAA2021}, %... for drones weighing under 55 pounds, recreational requires, between all other norms, fly at or below 400 feet in uncontrolled airspace. In the case 
%the Part 107 of the Code of Federal Regulations 
the rule for operating %Unmanned Aircraft Systems (UAS)
drones under $55lbs$ in the National Airspace System (NAS) is the {Code of Federal Regulations Part 107}~\cite{CFRpart107}, referred to as the Small UAV Rule. In section 39 of that rule specifies that no person may operate a small UAV over a human being unless that human being is directly participating in the operation, is located under a covered structure or inside a stationary vehicle that can provide reasonable protection from a falling small UAV or if the operation meets the requirements %of at least one of the operational categories specified in subpart D of this part.
of the subpart D of this part, which prescribes the eligibility and operation requirements for civil small UAV to operate over human beings.
This subpart %includes four categories, 
defines the requirements for different categories, but explicitly forbids all of them to operate in sustained flight over open-air assemblies of human beings unless otherwise authorized by the Administrator.}

\textcolor{black}{Following those rules, 
the Joint Authorities for Rulemaking on Unmanned Systems (JARUS) created a document of guidelines for aerial work operations called Specific Operations Risk Assessment (SORA)~\cite{s_2019}. It defines the risk as the combination of the frequency (probability) of an occurrence and its associated level of severity. Many categories of harm arise from any given occurrence, so, it defines three categories of harm:
\begin{itemize}
    \item Fatal injuries to third parties on the ground
    \item Fatal injuries to third parties in the air
    \item Damage to critical infrastructure
\end{itemize}
The consequence of an occurrence is designated as harm of some type. Thus, it defines two classes of risk: Ground Risk (GRC) and Air Risk (ARC). Each class has its own defined intrinsic risk. For UAS smaller than 1m of span, ground risks levels range from 1 to 8, and air risk from A to D. The calculus of those risks are based on UAV specifications, operation scenario and airspace. %In this work however, air risk will not be considered, and for ground, the intrinsic risk class estimated in the scope of this work is 5, where UAV fly over populated areas.
 %Once the intrinsic risk class is first obtained, different kinds of mitigations measures are determined to calculate the final or residual risk class. Then, tactical mitigations are applied during the conduct of the operation to mitigate any residual risk that may remain after the strategic mitigations have been applied. The Specific Assurance and Integrity Level (SAIL) is then determined using the operation concept and the consolidation of the final GRC and residual ARC into one of the six levels, where a high value represents an operation with high potential risk. For the assigned SAIL, the operator will be required to show compliance with %each 
 %the Operation Safety Obectives (OSO). Each OSO shall be met with a required Level of Robustness, which is a combination of the Level of Integrity and Assurance of each mitigation.
}
% https://www.youtube.com/watch?v=sq7wozIzXBM
\textcolor{black}{%In summary, in the frame of SORA, for the smaller category of UAV, the GRC depends on the operational scenario of controlling beyond visual line of sight and populated areas.
More specifically, the GRC relates to the unmitigated risk of a person being struck by the UAV (in case of loss of control) derived from the intended operation and the UAV lethal area, as shown in Table~\ref{tab:GRC}. As it can be seen, the most riskiest operation are over gatherings of people, what in regulations is mentioned as "assembly of people".
}

\begin{table}[H]
\caption{Intrinsic UAV Ground Risk Class for maximum characteristics dimension of 1m.}
\label{tab:GRC}
\begin{tabu} to 1.0\textwidth { X[1.5l]  X[0.5c] }
\toprule
\textbf{Operational Scenarios} & \multicolumn{1}{c}{\textbf{Class Level}} \\
\toprule
    VLOS/BVLOS over controlled ground area & 1 \\
\midrule
    VLOS in sparsely populated environment & 2 \\
\midrule
    BVLOS in sparsely populated environment & 3 \\
\midrule
    VLOS in populated environment & 4 \\
\midrule
    BVLOS in in populated environment & 5 \\
\midrule
    VLOS over gathering of people & 7 \\
\midrule
    BVLOS over gathering of people & 8 \\
\bottomrule    
\end{tabu}
\end{table}

\textcolor{black}{Following the SORA methodology, %it can be seen that urban operations of UAVs are risky because normally it can fall in the GRC 4 or 5. 
once the GRC is obtained, mitigation strategies should be applied to reduce the Final Ground Risk Class. This strategies are classified as:
\begin{itemize}
    \item M1 – means to reduce the number of people at risk.
    \item M2 – means to reduce the energy absorbed by the people of the ground upon impact.
    \item M3 – an Emergency Response Plan (ERP) is in place, operator validated and effective.
\end{itemize}
}

\textcolor{black}{%These mitigation strategies can be addressed with
%Depending on nationals laws, Autonomous Landing can be implemented as a M1 or M3 mitigation strategy. However, efectiveness 
%For an M3 mitigation strategy, %
The ERP should be defined by the applicant in the event of loss of control of the operation for emergency situations where the operation is in an unrecovarable state and in which the outcome of the situation highly relies on providence, could not be handled by a contingency procedure or when there is grave and imminent danger of fatalities. The ERP is expected to cover a plan to limit the escalating effect of a crash. %and the conditions to alert
In this context and depending on national laws, Autonomous Landing can be implemented as %M1 or 
M3 mitigation strategy.}

\subsection{Autonomous Landing}
Existing rotary-wing UAVs offer limited %autonomous landing 
Autonomous Landing capabilities, which is one of the main limitations for expanding the applications into urban populated areas. Although some sophisticated UAVs may be equipped with %security
\textcolor{black}{safety} mechanisms for emergency situations,  %most available drones are limited to
\textcolor{black}{%most only have} return-to-home policies in case of malfunction. 
the majority of currently available drones are constrained by return-to-home protocols in the event of malfunctions.} This normally involves %the UAV 
recording the take-off location, and, when communication is lost, returning to it autonomously and descending slowly. The device may also emit lights and sounds to alert unaware bystanders in the area. Nevertheless, such strategies do not warranty the safety of the landing missions, especially during a system failure or a battery shortage. More advanced drones could search for pre-known visual tags for landing \cite{garcia2002towards}, but this involves previously preparing the area, which is not ideal for most applications, especially those covering large areas, unstructured scenarios, or spaces with moving people or vehicles, as is common in urban areas.

Hence, %Autonomous Landing (AL)
%autonomous landing 
AL in urban environments is a complex and open research problem, due to the enormous variety of scenarios present in cities, where %not only the drone moves in an unknown unstructured scenario, maybe with a malfunction, but there are 
%the UAV navigates in an unknown unstructured scenario in different conditions, and there is also people, cars, animals, and other moving agents with unknown dynamics. %To acknowledge this problem, we can try to find a Safe Landing Zone (SLZ), which is defined as the best available area where the UAV may land without causing major damage. Then, 
the UAV operates within an unknown and unstructured environment under varying conditions, navigating alongside individuals, vehicles, animals, and other moving entities with unknown dynamics.
When a dangerous situation occurs, the drone must find a Safe Landing Zone (SLZ), which is defined as the best available space to land without causing any kind of damage, or %in other words, 
the area where the risk of material damage and hurting people is minimized. This has proven to be very challenging due to the diverse and changing nature of urban populated areas. %Proposing a solution that could be used out-of-the-box, using the components already installed in the UAV, could considerably broaden the real-life applications of them in urban areas.
Proposing a solution that can be readily deployed, leveraging the existing components integrated into the UAV, has the potential to greatly enhance their practical applications in urban settings.

In this regard, %%we consider
this work proposes the use of a %down-looking 
monocular camera attached to the UAV coupled with computer vision and %modern deep neural networks
% https://writingexplained.org/state-of-the-art-vs-state-of-the-art
% https://dictionary.cambridge.org/us/dictionary/english/state-of-the-art
% https://langeek.co/en/grammar/course/1618/state-of-the-art-vs-state-of-the-art
% https://ell.stackexchange.com/questions/272843/what-is-the-difference-between-state-of-the-art-and-state-of-the-art#:~:text=%22state%20of%20the%20art%22%20is,progress%20in%20a%20given%20field.&text=%22state%2Dof%2Dthe%2Dart%22%20is%20a%20compound,technology%20in%20a%20given%20field.
state-of-the-art %Machine Learning (ML) 
Deep Learning algorithms to analyze the scene and assess the risk of accidents that may produce significant material damage or, even worse, hurt human beings during emergency landing situations. 
%In this paper, we propose the use of a state-of-the-art 
For this, it will be used a semantic segmentation network based on Visual Transformers %, the SegFormer,
and trained on %the Semantic Drone Dataset \cite{semanticDroneDataset}.
UAV-obtained datasets, which are composed of aerial views of urban areas with labels at the pixel level of the most common classes for this context, %like 
such as people, cars, roads, vegetation, etc.
%Such a dataset is composed of aerial views in urban areas labeled for semantic segmentation, considering classes such as people, animals, grass, pavement, cars, vegetation, etc.
The semantic segmentation model will provide pixel-level classification of %monocular camera 
RGB images, which in turn will be used to %assess the risk 
\textcolor{black}{create a risk map assessing the risk} of material damage or human accidents in the areas beneath the drone. This risk map can be used later to perform further analysis and help the UAV with the decision-making process to minimize the risk of accidents, making the system more reliable and resilient. %, which in turn will allow to exploit the true potential of UAVs in smart cities.
%To narrow down the problem, n
\textcolor{black}{No 3D information of terrain will be accounted for (solely considerations that walls, doors, windows and other similar objects generate a specific risk level because of verticality), and suppose UAV work space is clear so there is no risk of air collision.}

\textcolor{black}{%This proposal will yield valuable information to an Uncrewed Aircraft System (UAS) using it to better understand the situation at ground in an urban area where people and motor vehicles among other things create a very dynamic scenario.
This proposal aims to furnish valuable insights to an Uncrewed Aircraft System (UAS) for a comprehensive understanding of the dynamic urban environment, particularly in areas where people and motor vehicles contribute to a multifaceted scenario.}

%The organization of this paper is as follows.
This work is organized as follows. Section~\ref{sec:related} contains a discussion on related work on %visual-based 
autonomous landing in general. Section~\ref{sec:method} %, a brief description of
describes the different modules that comprise the proposed strategy. % is presented. Later, we evaluate and discuss of the 
Experimental results are discussed and evaluated in Section~\ref{sec:experiment}. Finally, the conclusions and future work are presented in Section~\ref{sec:conclusions}.

\section{Related Works: Vision-Based Autonomous Landing}
\label{sec:related}
%Recent advances in the field of computer vision and machine learning have allowed the evolution of the autonomous landing task from only using sensors and radio navigation, such as GPS, to a more flexible and precise practice.
Since \cite{NIPS2012_c399862d} helped pave the way to Deep Learning, advances in the field of Computer Vision (CV) and Machine Learning (ML) allowed the evolution of Autonomous Landing (AL) tasks from simple state machines using ground beacons and radio telemetry sensors such as GPS, to a great diversity of techniques using only RGB cameras most of the time.

Some of the first works reported in the literature were devoted to the detection of previously known visual markers \cite{saripalliil_vision}, to avoid hazardous terrain \cite{johnsonil_vision_guided_landin_auton_helic_hazar_terrain}
or to use visual feedback to provide additional awareness to UAVs. These methods were based on extracting and then tracking traditional features from images obtained from a monocular camera as done in \cite{theodore2005full,kendoul2008adaptive}. However, vision was used mainly as an assistance system.

Furthermore, works based on people detection, such as \cite{rabaud2006counting,chan08_privac}, allowed %incorporating 
the incorporation of a more safety-oriented approach to autonomous landing, which was achieved by detecting people to avoid human accidents. Using basic computational vision tools, it is possible to find areas to land in simple and controlled scenarios. However, those tools become useless when the real environments in populated areas become too complex. The main troubles presented in previous works are not generalizing enough the features like heads of people, not accounting for changes in perspective, dealing with occlusion and scene movement, or restricting themselves to a handful of scenarios. 

Subsequent works like \cite{gonzaleztrejo2020lightweight} started focusing on improving people detection, being people the most critical factor during autonomous landing missions, by obtaining density maps from the people location directly from the images, to get people count and position data in crowded scenes. Those density maps were extracted using classical tools as Probability Density Function using FAST, from videos taken from airborne cameras above the crowds. However, this approach still fails with perspective variations.

Later on, proposals on the use of machine learning tools such as Convolutional Neural Networks (CNN) flourished, as these tools were capable of generalizing a wide variety of environments and situations as shown in \cite{gonzalez-trejo20_dense_crowd_detec_surveil_drones_densit_maps}. However, few works have been destined for determining UAVs' safe landing zones. 

Works such as \cite{liu19_geomet_physic_const_drone_based,li2018csrnet,Tzelepi2017HumanCrowd,tzelepi2019graph} use a Deep Neural Network (DNN) to try to find landing zones free of people. In \cite{tzelepi19_graph_embed_convol_neural_networ} a lightweight network is used to overestimate the density map of the crowd to prevent the drone from landing near any person. In addition, \cite{liu19_geomet_physic_const_drone_based} uses the pitch and altitude information provided by the drone to feed it to the DNN as an additional channel called the perspective map, from which a density map drawn on the head's plane is generated. This plane, which is where the heads are in average in the real world, prevents the system from underestimating people and helps to decide where to land.
\cite{gonzalez-trejo20_dense_crowd_detec_surveil_drones_densit_maps,gonzalez-trejo21_light_densit_map_archit_uavs} use lightweight networks to infer density maps based on a Bayesian-Loss network that is very suitable for execution in embedded systems. Then, in \cite{GonzalezRAL}, using the density map, the authors find areas devoid of people suitable for landing, and track them with Kalman filters coupled with the Hungarian algorithm.%, but they do not determine how to choose a SLZ to actually land. 

However, most proposals are hard to validate in real scenarios. Thus, \cite{tovanche2022vr} %also 
proposes a framework for safe real-time and a thorough evaluation of vision-based %autonomous landing 
AL in populated scenarios, using photo-realistic virtual environments and physics-based simulation. Then, software/hardware-in-the-loop can be used to test the performance of the algorithms beforehand. The final validation stage consists of a robot-in-the-loop evaluation strategy. To this end, a real drone must perform autonomous landing maneuvers in real-time, with an avatar drone in a virtual environment mimicking its behavior, while the detection algorithms run in the virtual environment (virtual reality to the robot). The authors also propose different metrics to quantify the performance of landing strategies, establishing a baseline for comparison with future work on this challenging task and analyze them through several randomized iterations.

Nevertheless, the previous 
%Unfortunately, these 
works are restricted to avoid accidents involving humans and do not account for potential material damage or indirect accidents involving people. %In addition, they usually assume that the terrain beneath is mainly a flat surface.
%In recent years, the progress on Machine Learning (ML) algorithms has accelerated, and thus the possibilities to rapidly improve the solutions to more complex research problems. 
\cite{symeonidis2021vision} proposes implementations of state-of-the-art techniques and tools in the search for SLZ for UAVs. In their article, the authors propose the use of different modules %in a pipeline 
to create an annotated surroundings map, with landing sites and no-fly zones around people locations, and continuously check for paths towards SLZ. %However, this algorithm cannot ensure finding a path to a SLZ in more complex scenarios, i.e., in the case of a very crowded area. 
\textcolor{black}{The approach considers finding people and different kinds of risks while also checking for a flat safe landing zone. However, the design of the system is focused in a sparse and less dynamic scenarios.}
Finally, \cite{chen2022robust} uses a binocular-LiDAR to find flat areas to land with a network doing semantic segmentation simultaneously. This double function helps the system understand the morphology and semantic features of the terrain, selecting SLZ in complex environments with very high accuracy. %Nevertheless, this proposal does not consider the dynamic scenarios people can create.
\textcolor{black}{%Although t
This method is good to give terrain context to the UAV, but people and other urban agents are not accounted for.}

\textcolor{orange}{Although some advances can be found in the recent literature, autonomous landing in dynamic urban unstructured environments continues to be a very complex open problem, where a lot of research effort is still required in order to have more reliable solutions.}
In this regard, the present work proposes a vision-based strategy for risk assessment in the context of autonomous landing in complex unstructured urban scenarios. The strategy consists in using a semantic segmentation DNN based on modern visual transformers to provide context to the UAV about the area %beneath 
below it. This is done by identifying classes of terrain and objects that can be used in the decision-making processes to avoid human accidents and material damage. %Classes employed include people, animals, cars, bicycles, grass, pavement, vegetation, etc. 
Furthermore, the inferred image from semantic segmentation is then mapped to a heat map that assesses the potential risk of accidents involving people, directly or indirectly, as well as potential material damage to expensive infrastructure or obstacles such as cars or to the drone itself. The proposed strategy can be easily used further as a framework for smart decision-making during emergency landing protocols. %, by coupling it with multi-objective optimization techniques, motion planning, feature tracking, etc. 
\textcolor{black}{The risk map can be coupled with multi-objective optimization techniques to take desitions on where to land, prioritizing the zones with less risk, even it could be elevated. }
The strategy is validated in various %cases of study
\textcolor{black}{%study cases
case studies}, showcasing the huge potential of %semantic segmentation-based techniques to assess risk 
this proposal to assess risk confidently in autonomous landing in urban complex unstructured scenarios.

\section{General Approach}
\label{sec:method}
%In this section, we introduce the modules and tools employed for the proposed risk assessment approach for autonomous landing.
The proposed risk assessment approach for AL %leverage the use of a couple of techniques
is composed of %two techniques presented next.
two stages: Semantic Segmentation and Clustering by Risk Level.

\subsection{Semantic Segmentation}
\label{sub:segmentacion}
%vvv
%Semantic segmentation (SS) is the task of classifying images in different categories at the pixel level. It was first proposed and developed in 2015 by Ronneberger et al. \cite{ronneberger2015u} as a tool for biomedical image segmentation, where it is needed to classify the tissues shown in screenings and signalize if diseases are present. The architecture proposed was called U-net and won the 2015 ISBI challenge. It is composed of a contracting path based on a typical Convolutional Neural Network (CNN), and an expansive symmetrical path. A final $1\times1$ convolution to determine the probability of each pixel being of a certain category.
%Main metrics used to evaluate it are the Dice Coefficient and Intersection over Union (IoU).
%\subsection{State of the Art on Semantic Segmentation}
%^^^
%In the context of our proposal, we define Semantic Segmentation (SS) as 
Semantic Segmentation (SS) is the task of classifying images into different categories at a pixel-wise level. This kind of classification is used because it produces better information on the context and location of the segmented objects. Historically, this task has improved greatly with the use of %Convolutional Neural Networks
CNNs, notably with the introduction of the first U-Net proposed %by 
in \cite{ronneberger2015u}.
%vvv
%\subsection{Visual Transformers}
%New approaches have been developed using transformers and other deep learning algorithms.
%The Transformer was proposed by Vaswani et al. \cite{vaswani17_atten_is_all_you_need} as a tool to process natural language, accounting for the most important context for the words being analyzed. Transformers use self-attention to allow key tokens to be amplified and less important ones to be diminished. The transformer was first applied to language translation, and its context-aware capabilities have made its incursion in different fields. More recently, SS networks have used transformers because they significantly improve accuracy.
%A transformer-based semantic segmentation network, such as UCFilTransNet from Li et al. \cite{li2018csrnet}, fuses multi-scale features information to improve segmentation accuracy. UCFilTransNet has the same number of parameters of a VGG16-based U-net, so it is a best option for modern-day applications. For this work, we have selected the Segformer \cite{xie2021segformer}, which has proven to be one of the best models for SS.
%^^^
More recently, the techniques developed for Natural Language Processing have generated great interest because they include diverse mechanisms %, such as multi-head attention, to 
in a model known as Transformer, that efficiently process vast and complex datasets. %that improves both accuracy and efficiency.
%the transformer proposed for natural language processing by \cite{vaswani17_atten_is_all_you_need} started to infuse into other fields thanks to the multihead attention mechanism that learns the dependencies between distant positions of the input and generates context for them.
%This mechanism proposed in the Transformer 
This model proposed by \cite{vaswani17_atten_is_all_you_need} started to infuse into other fields because it learns the dependencies between distant positions of the input and generates context.
These context-aware capabilities have allowed state-of-the-art models of SS to be very efficient and accurate. Models such as SegFormer~\cite{xie2021segformer} have a straightforward implementation that enables many downstream applications like the one presented in this paper, in addition to having a performance that makes it one of the best models for SS.

%To evaluate the performance of the semantic segmentation, works like \cite{ronneberger2015u}, \cite{xie2021segformer} use mean Intersection over Union (IoU). The usual metrics for classification, which include Accuracy, F1 Score (Dice Coefficient), and Balanced Accuracy, can also be used to give a broader understanding of the results.
Because Semantic Segmentation works at the pixel level, to measure performance of the SS models the community uses mean Intersection over Union (mIoU) and Dice Coefficient (DSC, also known as F1 Score) metrics, which are similarity coefficients between the predicted classes and the ground truth. %These metrics are defined as

%Mean 
Intersection over Union defined in Eq. %\ref{eq:mIoU} 
\ref{eq:IoU} serves as the fundamental metric to quantify the overlap between the predicted and ground truth areas. Higher values indicate better alignment between the predicted and actual regions, meaning a more accurate model.
%\begin{definition}
\begin{equation}
 %\label{eq:mIoU}
 \label{eq:IoU}
%mIoU = \frac{\left | A \cap B \right |}{\left | A \cup B \right |} = \frac{TP}{TP+FP+FN}
IoU = \frac{\left | A \cap B \right |}{\left | A \cup B \right |} = \frac{TP}{TP+FP+FN}
\end{equation}
%\end{definition}
So, mIoU is the average segmentation performance for all classes.
\begin{equation}
 \label{eq:mIoU}
    mIoU = \frac{1}{N} \sum_{i=1}^{N} IoU_i
\end{equation}
The DSC  defined in \ref{eq:F1} is the harmonic mean of precision and recall. In other words, it is the coefficient between detecting the target correctly and not detecting other classes.
\begin{equation}
    \label{eq:F1}
    F1_{score} = \frac{2 \times \left | A \cap B \right |}{\left | A \right | + \left | B \right |} = %2\times\frac{Precision \times Recall}{Precision + Recall}
    \frac{2 \times TP}{2 \times TP + FP + FN}
\end{equation}
%\noindent Displayed equations are centered and set on a separate line.
%\begin{equation}
%x + y = z
%\end{equation}

%\subsubsection{Sample Heading (Third Level)} Only two levels of headings should be numbered. Lower level headings remain unnumbered; they are formatted as run-in headings.
\subsubsection{Model}
\begin{figure}[b]
\centerline{\includegraphics[width=\columnwidth]{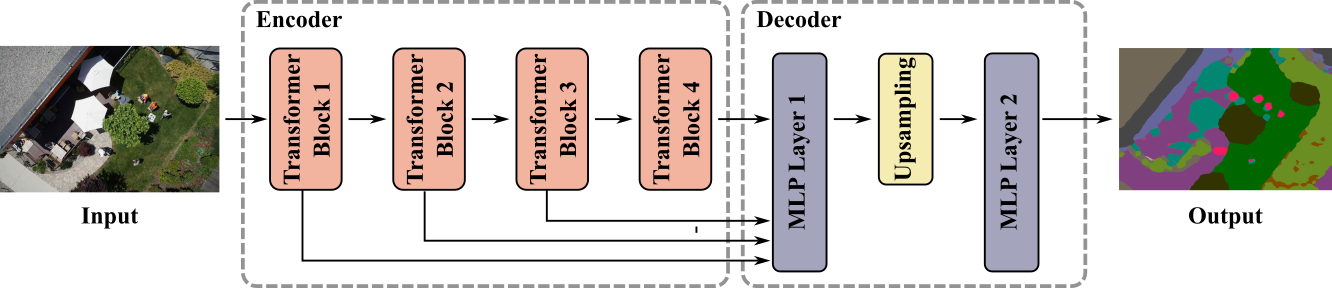}}
\caption{SegFormer architecture proposed by Xie \textit{et al.}~\cite{xie2021segformer}.}
\label{fig:segformer}
\end{figure}

%The model proposed for this task is the SegFormer. The SegFormer is an efficient SS framework that offers state-of-the-art results and flexibility, unifying visual transformers with lightweight multilayer perceptron (MLP) decoders. Its novel hierarchically structured Transformer encoder outputs multiscale features, as depicted in Fig.~\ref{fig:segformer}. Not using positional encoders avoids the interpolation of positional codes, which leads to increased performance when the testing resolution differs from that in training. Also, the MLP decoder aggregates information from different layers, combining local and global attention to render powerful representations. 
The implemented model for SS of images on this proposal is the SegFormer% network
. This sophisticated segmentation architecture that integrates Transformers and Multilayer Perceptron (MLP) blocks was proposed by %Xie \textit{et al.}~
\cite{xie2021segformer}. Its design offers the advantages of minimal parameters, fast training, and extensive computational capacity. The network utilizes an auto-encoder structure, as can be seen in Fig.~\ref{fig:segformer}. The encoder incorporates four transformer blocks that can produce features at multiple scales. The decoder employs an MLP to combine multi-scale features, complemented by an ``UpSample'' layer to restore the image resolution.

%Please try to avoid rasterized images for line-art diagrams and schemas. Whenever possible, use vector graphics instead (see Fig.~\ref{fig1}).

The simple and lightweight design of SegFormer offers a series of models with different sizes, from SegFormer-B0 to SegFormer-B5, allowing the flexibility of choosing the model best suited for the application needs. \textcolor{black}{The SegFormer chosen for this task was the one based on the backbone of MiT-B0}, which is the lightest and fastest version. This selection was based on the assumption that this model has to run in real-time on a computer embedded onboard the UAV at a sufficiently high rate to guarantee a safe landing.
%The SegFormer-B0 contains $3.7$ millions of parameters, being comparatively smaller and faster than other state-of-the-art models on semantic segmentation and producing a good \textit{mean IoU} of $0.53$ on three popular publicly available datasets.
%The SegFormer chosen for this task was the one based on the backbone of MiT-B5, which is the biggest and more accurate model

The model was implemented using the HuggingFace %~\cite{huggingFace}
library of the SegFormer~\cite{Rogge_2022}, which offers tools to build and test machine-learning architectures. %Its hub contains pre-trained models of all sizes. This proposal was based on the NVIDIA/MiT-B0 pre-trained on \textit{Imagenet-1k}.
The implemented model can receive inputs of ``any size'', reducing the preprocessing. The output is one-fourth of the size of the input, %but this scale does not produce any significant loss of information.%, and it can be used directly as is at the time of the application deployment.
so in training it was required to up-sample 4 times the output to compare it to ground truth. 
%The model can receive input of any size, and the logits at the output are input\_size/4, so it is recommended to upscale it. In this work, the interpolation function from torch.nn is already optimized for tensors given by the models of pytorch. The up scaling is only necessary for training, validation, and testing, to compare directly with ground truth, but an application using this model can directly use the output and make decisions directly on it.

%\subsection{Fast Segmentator}
%The last model tested is the following

% \subsection{Models Output}\label{Output}
% Semantic Segmentation is normally interpreted visually as a colored image, where each color represents a category. However, those categories are obtained as a function of probability of each pixel. In order of selection the best SLZ, it will be useful to use directly the probability of categories for each pixel.

%\paragraph{Sample Heading (Fourth Level)}
%The contribution should contain no more than four levels of headings. Table~\ref{tab1} gives a summary of all heading levels.

\subsubsection{Dataset}
A key element for training %deep neural networks 
DNN is the dataset. %In that sense, we have trained the %DNN model 
%SegFormer 
\textcolor{black}{To take advantage of the SegFormer, it was finetuned on the Semantic Drone Dataset %from Graz University of Technology
(SDD)~\cite{semanticDroneDataset}, a dataset designed to enhance the safety of autonomous drone flight and landing procedures through improved semantic comprehension of urban environments. It is publicly available and contains $400$ high-resolution aerial images of size $6000\times4000$ pixels ($24$Mpx) from bird’s eye view, acquired %from a drone 
at altitudes between $5$ to $30$ meters above ground.}

\begin{figure}[H]
\centering
%\begin{subfigure}{0.5\textwidth}
%    \includegraphics[width=0.325\columnwidth,angle=90,origin=c]{low_SSdata4.jpg}
%    \includegraphics[width=0.325\columnwidth,angle=90,origin=c]{low_SSdata3.jpg}
%\end{subfigure}
%\hfill
%\begin{subfigure}{0.5\textwidth}
%    \includegraphics[width=0.49\columnwidth]{low_SSdata2.jpg}
%    \includegraphics[width=0.49\columnwidth]{low_SSdata5.jpg}
%\end{subfigure}
\includegraphics[width=0.24\columnwidth]{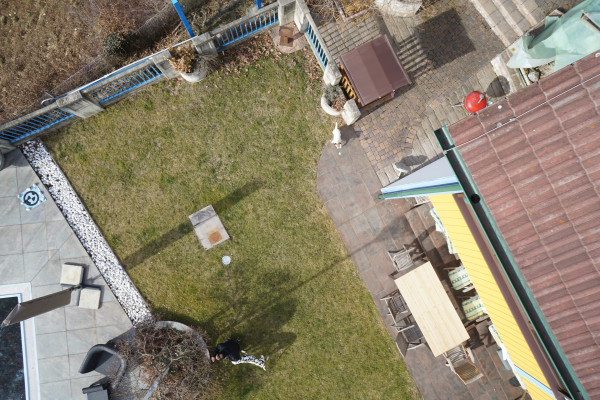}
\includegraphics[width=0.24\columnwidth]{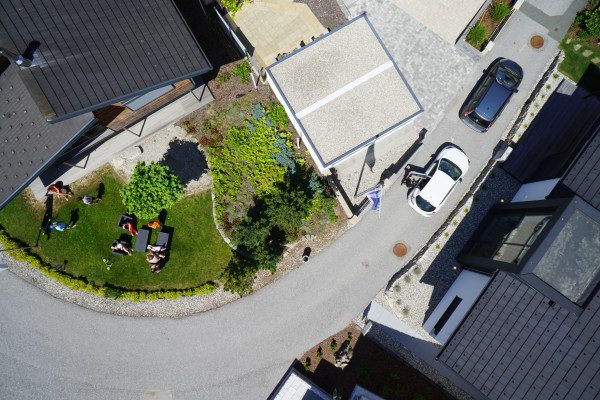}
\includegraphics[width=0.24\columnwidth]{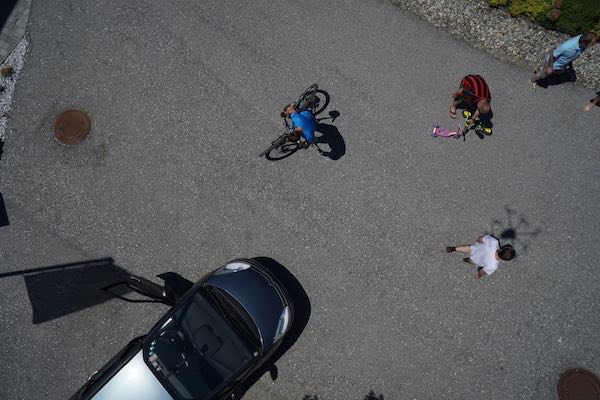}
\includegraphics[width=0.24\columnwidth]{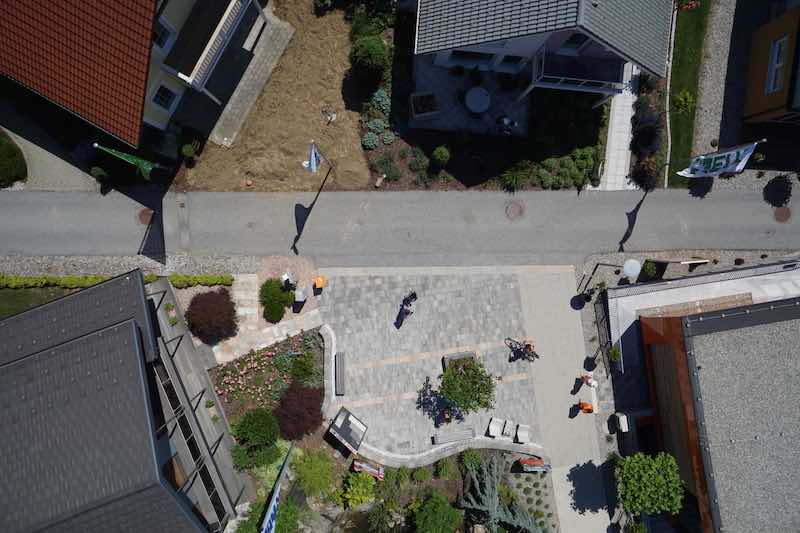}
\caption{Samples of aerial images in the Semantic Drone Dataset, captured in urban environments, using a down-looking camera mounted on a UAV. These images are labeled with several common classes found in urban areas, including grass, pavement, people, cars, vegetation, etc.}
\label{fig:dataset}
\end{figure}

Fig.~\ref{fig:dataset} depicts some samples of this dataset. Annotations are provided for SS and include %the 
$23$ of the most common classes from outdoor environments, including people, cars, vegetation, grass, pavement, water, animals, etc. 
%, as presented in Table~\ref{tab:classes}.
Table~\ref{tab:classes} presents a comprehensive overview of the dataset classes alongside their respective colors. Additionally, the proposed risk classification is displayed along with its corresponding color, as will be explained %later on.
in Subsection~\ref{sub:risk}

\begin{table}[h!t!b]
\caption{Classes used for semantic segmentation and their corresponding Risk Levels.}
\label{tab:classes}
\begin{center}
  % https://tex.stackexchange.com/questions/134381/dealing-with-very-long-table-split-into-columns
  %\begin{tabular}[t]{cc}   % Se cambió por minipage, como sea no era necesaria, pues hacer dos tablas sale lo mismo
  \begin{minipage}{0.49\textwidth}
  \begin{center}
    \begin{tabular}{ccc|cc}
    \toprule
    \multicolumn{3}{c}{\textbf{Semantic}} & \multicolumn{2}{c}{\textbf{Risk}} \\
    \multicolumn{3}{c}{\textbf{Segmentation}} & \multicolumn{2}{c}{\textbf{Level}} \\
    \midrule
    \textit{Class} & \textit{Label} & \textit{RGB} & \textit{Class} & \textit{RGB}\\
    \midrule
    0 & background & \tikz\draw[mycolor0, fill=mycolor0](-0.05,-0.05) rectangle (0.5,0.1); & \multirow{5}{*}{0}  & \multirow{5}{*}{\tikz\draw[mycolorR0,fill=mycolorR0](-0.05,-0.05) rectangle (0.5,0.1);} \\
    1 & dirt & \tikz\draw[mycolor1,fill=mycolor1] (-0.05,-0.05) rectangle (0.5,0.1); &  & \\%\multirow{4}{*}{0}  & \multirow{4}{*}{\tikz\draw[mycolorR0,fill=mycolorR0] (-0.05,-0.05) rectangle (0.5,0.1);} \\
    2 & grass & \tikz\draw[mycolor2,fill=mycolor2] (-0.05,-0.05) rectangle (0.5,0.1); &  & \\
    3 & gravel & \tikz\draw[mycolor3,fill=mycolor3] (-0.05,-0.05) rectangle (0.5,0.1); &  & \\
    4 & ar-marker & \tikz\draw[mycolor4,fill=mycolor4] (-0.05,-0.05) rectangle (0.5,0.1); &  & \\
    \midrule
    5 & paved-area & \tikz\draw[mycolor5,fill=mycolor5] (-0.05,-0.05) rectangle (0.5,0.1); & \multirow{2}{*}{1} & \multirow{2}{*}{\tikz\draw[mycolorR1,fill=mycolorR1] (-0.05,-0.05) rectangle (0.5,0.1);}\\
    6 & vegetation & \tikz\draw[mycolor6,fill=mycolor6] (-0.05,-0.05) rectangle (0.5,0.1); &   & \\
    \midrule
    7 & rocks & \tikz\draw[mycolor7,fill=mycolor7] (-0.05,-0.05) rectangle (0.5,0.1); & \multirow{5}{*}{2} & \multirow{5}{*}{\tikz\draw[mycolorR2,fill=mycolorR2] (-0.05,-0.05) rectangle (0.5,0.1);} \\
    8 & pool & \tikz\draw[mycolor8,fill=mycolor8] (-0.05,-0.05) rectangle (0.5,0.1); &  & \\
    9 & roof & \tikz\draw[mycolor9,fill=mycolor9] (-0.05,-0.05) rectangle (0.5,0.1); &  & \\
    10 & fence & \tikz\draw[mycolor10,fill=mycolor10] (-0.05,-0.05) rectangle (0.5,0.1); &  & \\
    11 & fence-pole & \tikz\draw[mycolor11,fill=mycolor11] (-0.05,-0.05) rectangle (0.5,0.1); &  & \\
    %\midrule   % Sale movida la tabla
    \bottomrule
    \end{tabular}
  \end{center}
  \end{minipage}
  \begin{minipage}{0.49\textwidth}
  \begin{center}
    \begin{tabular}{ccc|cc}
    \toprule
    \multicolumn{3}{c}{\textbf{Semantic}} & \multicolumn{2}{c}{\textbf{Risk}} \\
    \multicolumn{3}{c}{\textbf{Segmentation}} & \multicolumn{2}{c}{\textbf{Level}} \\
    \midrule
    \textit{Class} & \textit{Label} & \textit{RGB} & \textit{Class} & \textit{RGB}\\
    \midrule
    12 & water & \tikz\draw[mycolor12,fill=mycolor12] (-0.05,-0.05) rectangle (0.5,0.1); & \multirow{8}{*}{3} & \multirow{8}{*}{\tikz\draw[mycolorR3,fill=mycolorR3] (-0.05,-0.05) rectangle (0.5,0.1);}\\
    13 & wall & \tikz\draw[mycolor13,fill=mycolor13] (-0.05,-0.05) rectangle (0.5,0.1); &  & \\
    14 & window & \tikz\draw[mycolor14,fill=mycolor14] (-0.05,-0.05) rectangle (0.5,0.1); &  & \\
    15 & door & \tikz\draw[mycolor15,fill=mycolor15] (-0.05,-0.05) rectangle (0.5,0.1); &  & \\
    16 & bicycle & \tikz\draw[mycolor16,fill=mycolor16] (-0.05,-0.05) rectangle (0.5,0.1); &  & \\
    17 & tree & \tikz\draw[mycolor17,fill=mycolor17] (-0.05,-0.05) rectangle (0.5,0.1); &  & \\
    18 & bald-tree & \tikz\draw[mycolor18,fill=mycolor18] (-0.05,-0.05) rectangle (0.5,0.1); &  &  \\
    19 & obstacle & \tikz\draw[mycolor19,fill=mycolor19] (-0.05,-0.05) rectangle (0.5,0.1); &   &  \\
    \midrule
    20 & dog & \tikz\draw[mycolor20,fill=mycolor20] (-0.05,-0.05) rectangle (0.5,0.1); & \multirow{3}{*}{4} & \multirow{3}{*}{\tikz\draw[mycolorR4,fill=mycolorR4] (-0.05,-0.05) rectangle (0.5,0.1);}\\
    21 & car & \tikz\draw[mycolor21,fill=mycolor21] (-0.05,-0.05) rectangle (0.5,0.1); &  & \\
    22 & conflicting & \tikz\draw[mycolor22,fill=mycolor22] (-0.05,-0.05) rectangle (0.5,0.1); &  & \\
    \midrule
    23 & person & \tikz\draw[mycolor23,fill=mycolor23] (-0.05,-0.05) rectangle (0.5,0.1); & 5  & \tikz\draw[mycolorR5,fill=mycolorR5] (-0.05,-0.05) rectangle (0.5,0.1);\\
    \bottomrule
    \end{tabular}
  %\end{tabular}
  \end{center}
  \end{minipage}
\end{center}
\end{table}

%The Semantic Drone Dataset is 
To finetune the SegFormer, the SDD was divided as follows: $80\%$ for training, $10\%$ for validation, and $10\%$ for testing. Then, to increase the diversity of the data, a data augmentation technique %is 
was used to increase four times the number of images. In order to generalize aerial footage taken from a UAV, the transformations applied to augmented images included only brightness and contrast change, rotations, flips and random crops (transformations likely to happen to footage get with UAVs).

%\subsection{Testing Different Semantic Segmentation Networks}
%Starting with U-net based on the classifier VGG16.
%A VGG16 model pre-trained on the ImageNet dataset has been used as an encoder for the network. Decoder path has been extended from the last layer of the pre-trained model, and it is concatenated to the consecutive convolution blocks. Its input shape is (512, 512, 3). Trained over 20 epochs.
%The second model we tested is the SegFormer, a transformer-based semantic segmentator with high precision.
%Finally, we will use the FastSegmentator, which is a well-balanced model especially designed for embedded systems.

%\subsection{Implementation}
%We evaluated three configurations of the dataset for each model, and made a comparison of the main metrics described.

%\subsection{U-net based on VGG16}
%This is the fully structure of the network:

%\subsection{Risk Assessment}
\subsection{Clustering by Risk Level}
\label{sub:risk}
%In the context of autonomous landing in complex urban scenarios, semantic segmentation with a large number of classes may lead to errors, due to the difficulty of the task itself. 
A large number of classes on the semantic segmentation model may provide unnecessary information and affect negatively its metrics.
%Furthermore, it provides much more information than is needed, which may be expensive to compute in real-time by a processor embedded in the UAV. Consequently, we propose using semantic segmentation information solely as a stepping stone toward our ultimate goal of determining different levels of risk. 
Besides, to assess the risk of an scene using all classes individually could be more challenging, so it is useful to group classes with similar risk profile.

Following the SORA, it was decided to define $6$ different levels of risk, 
based on possible human injuries (which establishes the highest risk), material damage, preservation of the integrity of the drone and the identification of ideal zones for landing, 
being 0 the most ideal area to land, and 5 %where people are, 
the riskiest, as defined in Table~\ref{tab:risklevel_def}. 
This parameter was selected %by methodology 
to reflect the intrinsic risk class proposed by SORA but it could be easily adjusted for different applications. Depending on the risk associated, the segmented classes are classified as observed in Table~\ref{tab:classes}.

%\begin{itemize}
%    \item Level $5$ : This level represents the maximum risk and considers the direct risk of hurting people.
%    \item Level $4$: This level comprises indirect risk of hurting people, direct risk of hurting fauna,  and conflicting regions where there is uncertainty about the presence of people in the area.
%    \item Level $3$: This level includes important material damage, the imminent risk of losing or critically damaging the drone, and the moderate risk of indirectly hurting people. It includes the classes water, tree, window, wall, among others.
%    \item Level 2: Moderate risk of loosing or damaging the drone, along with low risk of material damage.
%    \item Level 1: Low level of material damage or to the drone itself.
%    \item Level 0: Ideal landing zones, including grass, dirt, gravel, and predefined markers.
%\end{itemize}
%%===========================
% NOTA: Seleccionar la tabla o el texto
%%===========================
\begin{table}[H]
\caption{Risk level definition.}
\label{tab:risklevel_def}
\begin{tabu} to 1.0\textwidth { X[0.5c]  X[1.7l] }
\toprule
\textbf{Risk Level} & \multicolumn{1}{c}{\textbf{Definition}} \\
\toprule
    0 &  Ideal landing zones, including grass, dirt, gravel, and predefined markers. \\
\midrule
    1 & Low level of material damage or damage to the UAV itself.\\
\midrule
    2 &  Moderate risk of loosing or damaging the UAV, along with low risk of material damage.\\
\midrule
    3 & This level includes important material damage, the imminent risk of losing or critically damaging the drone, and the moderate risk of indirectly hurting people. It includes the classes water, tree, window, wall, among others.\\
\midrule
    4 & This level comprises indirect risk of hurting people, direct risk of hurting fauna,  and conflicting regions where there is uncertainty about the presence of people in the area.\\
\midrule
    5 & This level represents the maximum risk and considers the direct risk of hurting people.\\
\bottomrule    
\end{tabu}
\end{table}

% Metrics given by the SegFormer for this particular dataset are in the model metrics, but this can be improved if the classes are grouped by the risk this particular category poses in terms of a flying object, being people the highest risk. \ref{tab1} show the risk class assigned to each one of the original categories. Categories such as dirt, grass, gravel, etc. are considered the lowest risk because those are areas typically used for take-off and landing of UAV. Other categories in the spectrum can be grouped for characteristics, such as wall, window, door, because these normally are in a vertical plane where is not possible to land the UAV. Obstacle was assigned a medium risk because the wide field of objects grouped by this class, so it was fixed at the middle.
%The risk given to each class is shown in the table

\begin{figure}[t!]
%\label{fig:process_example}
% \label without proper reference on input line 423. See the caption package documentation for explanation. -- https://tex.stackexchange.com/questions/279701/i-cant-figure-out-why-my-ref-is-displaying-question-marks-instead-of-a-figure-n
%\centering
%\begin{subfigure}{0.5\textwidth}
%    \includegraphics[width=0.45\textwidth]{106low_original.jpg}
%    \includegraphics[width=0.45\textwidth]{106low_ref.jpg}
%\end{subfigure}
%\hfill{}
%\begin{subfigure}{0.5\textwidth}
%    \includegraphics[width=0.45\textwidth]{106low_pred.jpg}
%    \includegraphics[width=0.45\textwidth]{106low_pred_risk_map.jpg}
%    \label{fig:process}
%\end{subfigure}
%\caption{The process of risk assessment, inference and training: top left picture shows the original image, top right image is the ground truth reference for training, bottom left image shows the result obtained from the semantic segmentation model (colored for visual interpretation), and  the bottom right image presents the mapping to risk levels, where red regions are considered the most dangerous and blue regions are the safest.}
%\label{fig:riskExample}
    \includegraphics[width=0.24\textwidth]{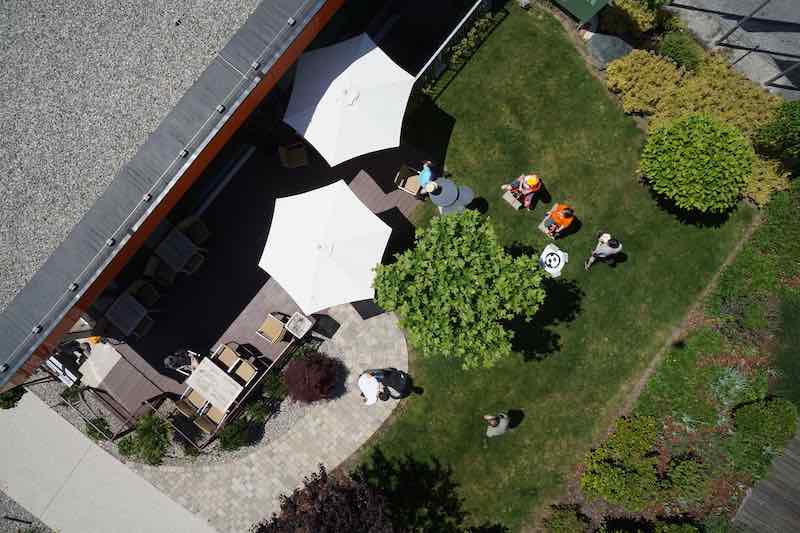}
    \includegraphics[width=0.24\textwidth]{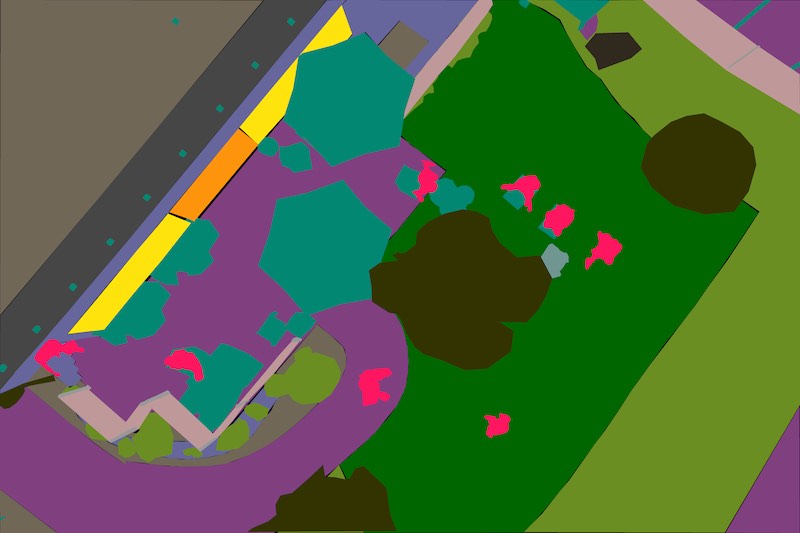}
    \includegraphics[width=0.24\textwidth]{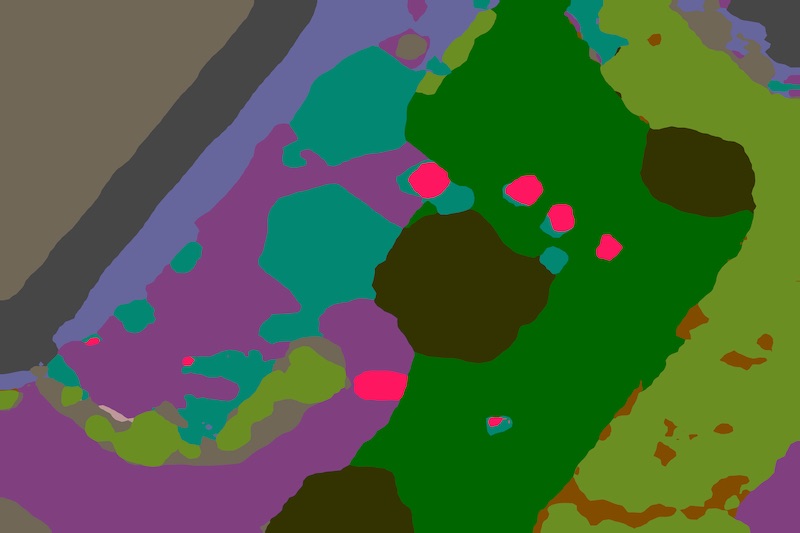}
    \includegraphics[width=0.24\textwidth]{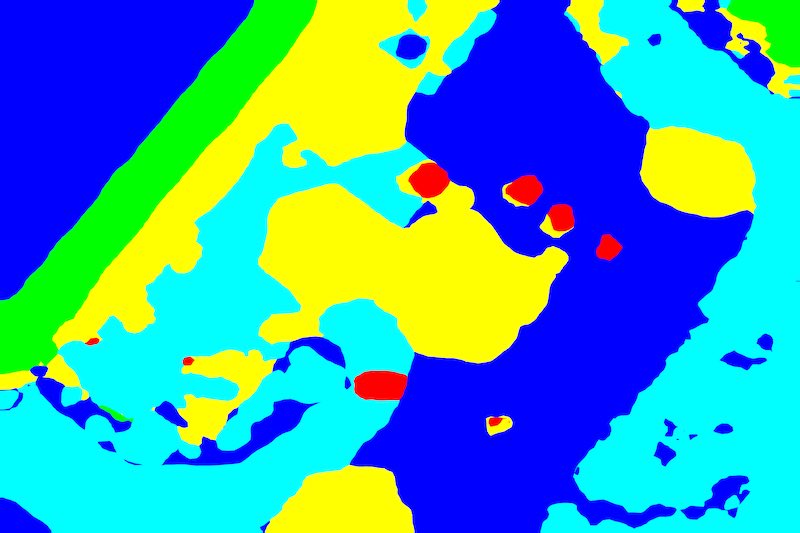}
\caption{The process of inference and risk assessment: first image shows the input image, the second is the ground truth reference of semantic segmentation, the third picture shows the prediction inferred from model (colored for visual interpretation), and last image presents the mapping to risk levels, where red regions are the riskiest and blue the safest. Reference for colors in Table~\ref{tab:classes}.}
\label{fig:process_example}
\end{figure}

In this work, conversion from semantic segmentation classes to risk levels is performed after inference%. %That is, using the trained model with the original $23$ categories, the predicted classes are converted to the new risk categories from the inferred output. Alternatively, a new semantic segmentation model could be trained to directly classify risk levels. 
, so the model can learn the characteristics of the same kind of classes and not lose generalization.
Fig.~\ref{fig:process_example} shows an example of the different stages of the risk assessment strategy. %In the top left figure, we can see the original image, in the top right the ground truth labels used for the segmentation training, then in the left bottom we have the inferred semantic segmentation image, and finally, in the bottom right image we can appreciate the mapping to the risk levels as some sort of heat map, where the hottest regions (red) represent greater risk, while the colder ones (blue) represent less risk.

%\subsection{Metrics}
%To evaluate the experimental results, the usual metrics for classification and semantic segmentation are used, that is, Accuracy, Intersection over Union (IoU), F1 Score (Dice Coefficient) and Balanced Accuracy. %Metrics were measured per image on the label output, but per class for the risk-converted output.

\section{Experimental Validation}
\label{sec:experiment}

\subsection{Experimental Setup}
\label{sub:exp_setup}
Development\textcolor{black}{, training the model} and testing were carried out with Python virtual environments on two desktop computers running on Ubuntu 20.04 equipped with processors Intel(R) Core(TM) i9-9900K and i7-8700K CPU and graphic cards NVIDIA GeForce RTX 2080 Ti and GTX 1080 Ti% GPU cards. %Ubuntu 20.04 is installed as Operating System of the custom computer. Python was chosen as the language in a virtual environment.
\textcolor{black}{. %Testing the frame rate at which the model and the risk assessment proposal run at an embedded system was done on an NVIDIA Jetson AGX Xavier (32 GB). This was only to obtain the minimum FPS produced by running the model. The result was 14 FPS.
An NVIDIA Jetson AGX Xavies (32 GB) was used only to obtain the minimum frame rate at which the SS model and the risk assessment proposal will run.}

The model was developed using Pytorch libraries and the SegFormer implementation by the HuggingFace community. %NVIDIA has six different variations of the model available on the HuggingFace Hub. These models are already pre-trained in \textit{Imagenet-1k}, which contains a large number of images and the $1000$ most basic categories.
%The SegFormer chosen for this task was the one based on the backbone of MiT-B0, which is the lightest and fastest version. This selection was based on the assumption that this model has to run in real-time on a computer embedded onboard the UAV at a sufficiently high rate to guarantee a safe landing.
%The SegFormer-B0 contains $3.7$ millions of parameters, being comparatively smaller and faster than other state-of-the-art models on semantic segmentation and producing a good \textit{mean IoU} %from almost $0.37$ to 
%of $0.53$ on three popular publicly available datasets.%ADE20K, Cityscapes and COCOStuff datasets.
At the HuggingFace Hub, NVIDA offers a MiT-B0 SegFormer encoder pretrained on \textit{Imagenet-1k} %which will be fine tuned on the Semantic Drone Dataset.
which was finetuned on the SDD.
%NVIDIA offers a MiT-B0 already pre-trained on \textit{Imagenet-1k} at the HuggingFace Hub.

\subsection{Experimental Results}
Different model trainings were conducted to assess the best configuration of the model for the application. The training dataset with data augmentation and all its classes were fed into the model. After $10$ epochs, the validation metrics were still improving, so the model was trained for further epochs. At $30$ epochs, the metrics started to plateau, and therefore training was stopped to avoid overfitting.
\textcolor{black}{After integrating the SS model to the risk assessment pipeline, the system produced a minimum frame rate of 14 FPS at the Jetson Xavier, showing its viability to integrate the proposal to the flight system of certain UAVs.}
In terms of semantic segmentation, the results given by the model trained match the reported in the literature.

\begin{table}[b!]
\caption{Metrics obtained from experiments.}
\begin{center}
\begin{tabular}{ccccc}
\toprule
\textbf{Experiment} & \textbf{\textit{Acc.}} & \textbf{\textit{Mean IoU}} & \textbf{\textit{F1 Score}} & \textbf{\textit{Bal Acc.}}\\
\midrule
%Class Segmentation & 0.8837 & 0.3946 & 0.4654 & 0.5397 \\
Class Segmentation & 0.8831 & 0.4138 & 0.4888 & 0.5754 \\
%\hline
%Risk Levels & 0.9031 & 0.5130 & 0.6000 & 0.6361 \\
Risk Levels & 0.8976 & 0.5811 & 0.6725 & 0.7110 \\
\bottomrule
\end{tabular}
\label{tab:testmetrics}
\end{center}
\end{table}

%\begin{table}[bh!]
%\begin{table}[h!]
%\caption{\textcolor{black}{Per class metrics of risk levels}}
%\begin{center}
%\begin{tabular}{ccccc}
%\toprule
%\textbf{Risk Class} & \textbf{\textit{Acc.}} & \textbf{\textit{Mean IoU}} & %\textbf{\textit{F1 Score}} & \textbf{\textit{Bal Acc.}}\\
%\midrule
%0 & 0.941664 & 0.822282 & 0.902475 & 0.923288 \\
%1 & 0.927699 & 0.874056	& 0.932796 & 0.927199 \\
%2 & 0.978853 & 0.678554 & 0.808498 & 0.923305 \\
%3 & 0.953360 & 0.596212	& 0.747033 & 0.883382 \\
%4 & 0.999042 & 0.849466	& 0.918607 & 0.940842 \\
%5 & 0.994712 & 0.572178	& 0.727880 & 0.845290 \\
%\bottomrule
%\end{tabular}
%\label{tab:testmetrics}
%\end{center}
%\end{table}

%\textcolor{black}{tabla con los hiperparametros utilizados}

% \begin{figure}[t]
% \centerline{\includegraphics[width=0.5\textwidth]{Accuracy.eps}}
% \caption{Training and Validation Accuracy for experiment 1}
% \label{X1_accuracy}
% \end{figure}

%\begin{figure}[t]
%\centerline{\includegraphics[width=0.5\textwidth]{TrainingAccuracy.eps}}
%\caption{poner el pie de foto.}
%\label{img_training_acc}
%\end{figure}

% \begin{figure}[t]
% \centerline{\includegraphics[width=0.5\textwidth]{Loss.eps}}
% \caption{Training and Validation Loss for experiment 1}
% \label{X1_Loss}
% \end{figure}

%\begin{figure}[t]
%\centerline{\includegraphics[width=0.5\textwidth]{TrainingLoss.eps}}
%\caption{poner el pie de foto.}
%\label{img_training_acc}
%\end{figure}

\begin{figure}[h]
\centerline{\includegraphics[width=0.8\columnwidth]{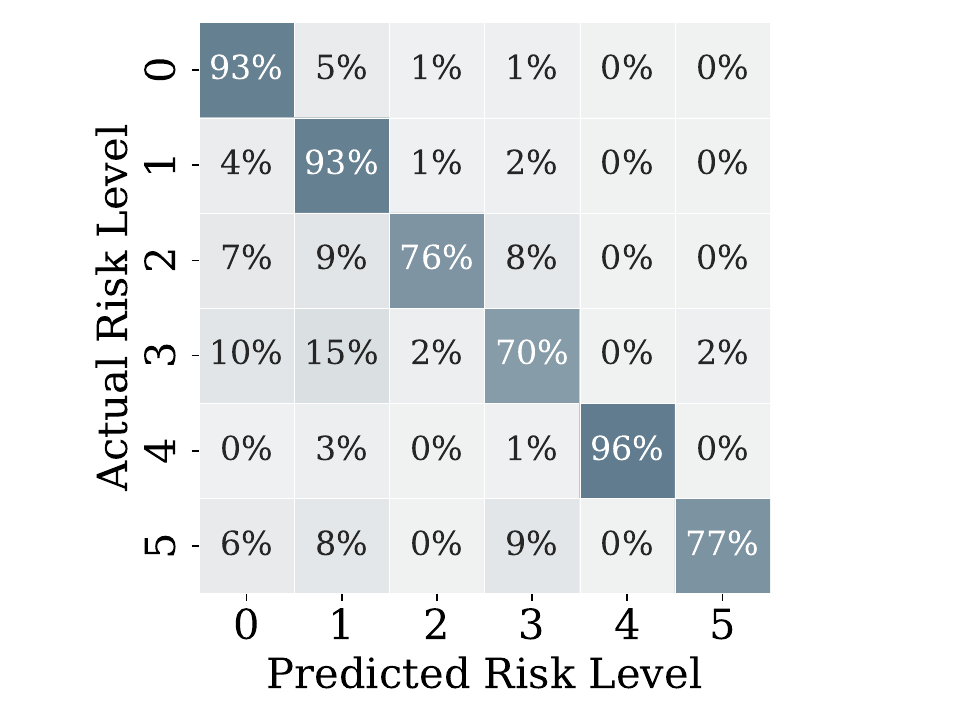}}
\caption{Confusion matrix of the testing set on the six risk levels.}
\label{fig:confusionmatrix}
\end{figure}

%For testing this first experiment, evaluations were done on the test set given the results shown in table ....
The evaluation was carried out in the test set, producing the results shown in Table \ref{tab:testmetrics}. 
%It can be seen that reducing and grouping the categories, even after the model produced the $24$ label inference, every metric is improved.
It can be seen that grouping categories by risk level implies reducing the number of classes, which also helps to improve the metrics obtained, as seen in Table \ref{tab:testmetrics}. This suggests that risk-level conversion is able to cope with inaccuracies in semantic segmentation, which, in turn, %also 
suggests that the process of making decisions about safe landings based on risk levels %might be 
is more robust than making %decisions 
the decision directly on semantic segmentation. %Also, note that the mapping to risk levels was done after inference, which means that probably training the model directly on the risk-level classes will improve metrics. 
Fig.~\ref{fig:confusionmatrix} shows the confusion matrix when the prediction is transformed to risk levels. Risk levels $0$ and $1$, which could be considered desirable %landing locations
locations to land, had a correct labeling above $90\%$. The same for level $4$. However, in the case of risk levels %1, 2 
$2$, $3$ and $5$, accuracy falls below $80\%$, something not so desirable, especially for the most critical level $5$.
%For level 5 the model predicted a $10\%$ to be from levels 0, 1 and 3, being the lower levels crucial to be differentiated from the higher levels.
Most classes and by extension, risk levels, were normally predicted to be of the classes with lower risk for up to $15\%$. Those risk levels contain the most common and extensive classes in the datasets, i.e. grass, dirt, paved-area, etc. %This only points to classes not balanced in the dataset and that the model requires more training on a more diverse dataset.
\textcolor{black}{The reason to this is that normally in the field of semantic segmentation, "background" classes as those mentioned earlier, compose the major part of the images due to its nature, technically unbalancing the dataset, in an unavoidable way. Even more, those classes encircle all the others, this being the main reason why all other classes have a considerable percentage being predicted as those.}
%However, confusion with the high risk level classes should be addressed separately to prevent the UAV from causing an accident.
Thus, it is preferable to have more false positives on the high risk level classes than false negatives. In this regard, one strategy is to dilate the high levels areas, so the resulting prediction could be safer for critical classes.

%\begin{theorem}
%This is a sample theorem. The run-in heading is set in bold, while the following text appears in italics. Definitions, lemmas, propositions, and corollaries are styled the same way.
%\end{theorem}
%% the environments 'definition', 'lemma', 'proposition', 'corollary', 'remark', and 'example' are defined in the LLNCS documentclass as well.

%\begin{proof}
%Proofs, examples, and remarks have the initial word in italics, while the following text appears in normal font.
%\end{proof}

\begin{figure*}[ht!]
\centering
\begin{subfigure}{\textwidth}
\centerline{a) \includegraphics[width=0.23\textwidth]{106low_original.jpg}
\includegraphics[width=0.23\textwidth]{106low_ref.jpg}
\includegraphics[width=0.23\textwidth]{106low_pred.jpg}
\includegraphics[width=0.23\textwidth]{106low_pred_risk_map.jpg}}
\end{subfigure}
\hfill
\hfill
\begin{subfigure}{\textwidth}
\centerline{b) \includegraphics[width=0.23\textwidth]{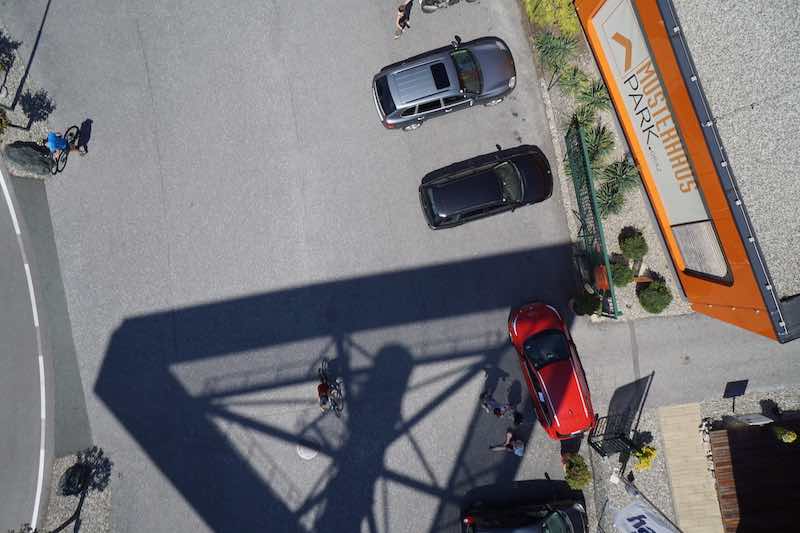}
\includegraphics[width=0.23\textwidth]{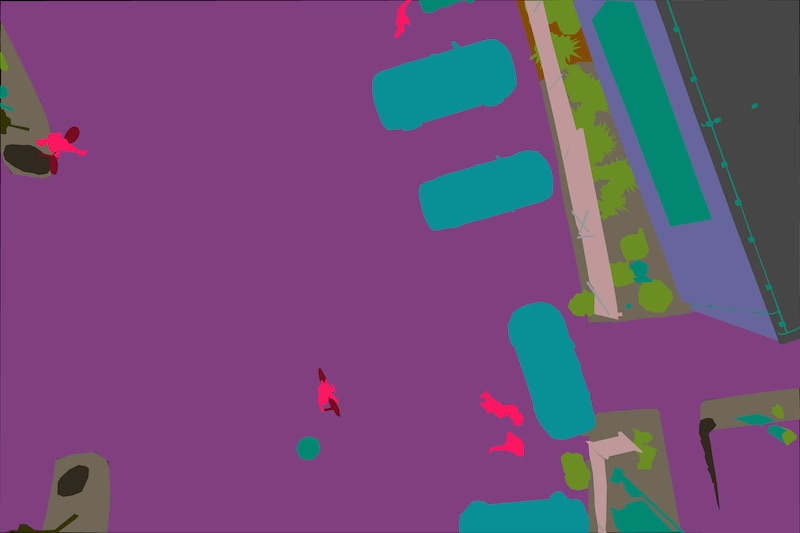}
\includegraphics[width=0.23\textwidth]{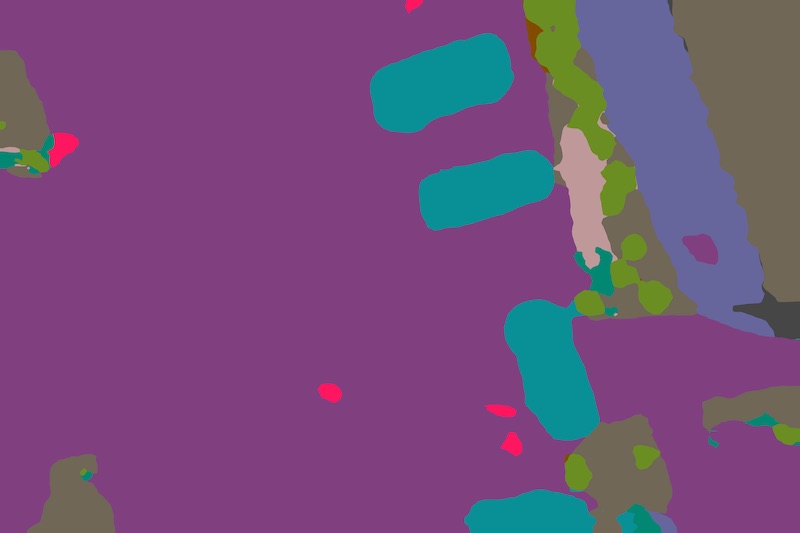}
\includegraphics[width=0.23\textwidth]{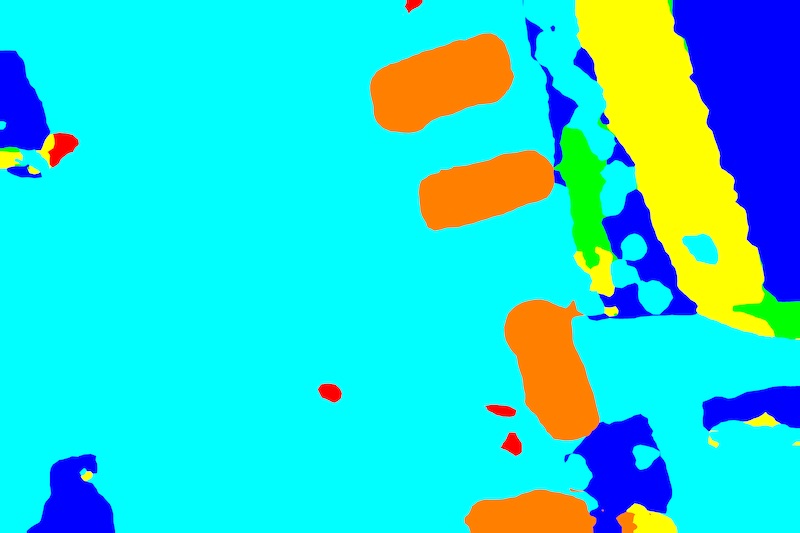}}
\end{subfigure}
\hfill
\hfill
\begin{subfigure}{\textwidth}
\centerline{c) \includegraphics[width=0.23\textwidth]{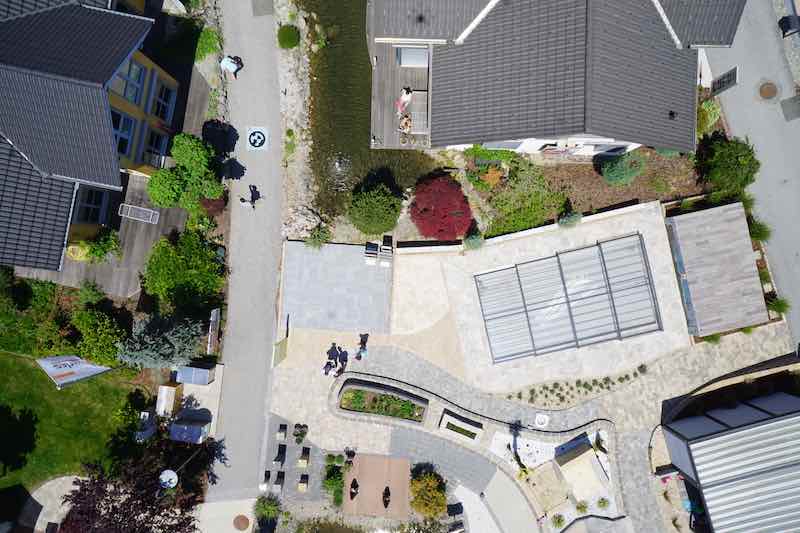}
\includegraphics[width=0.23\textwidth]{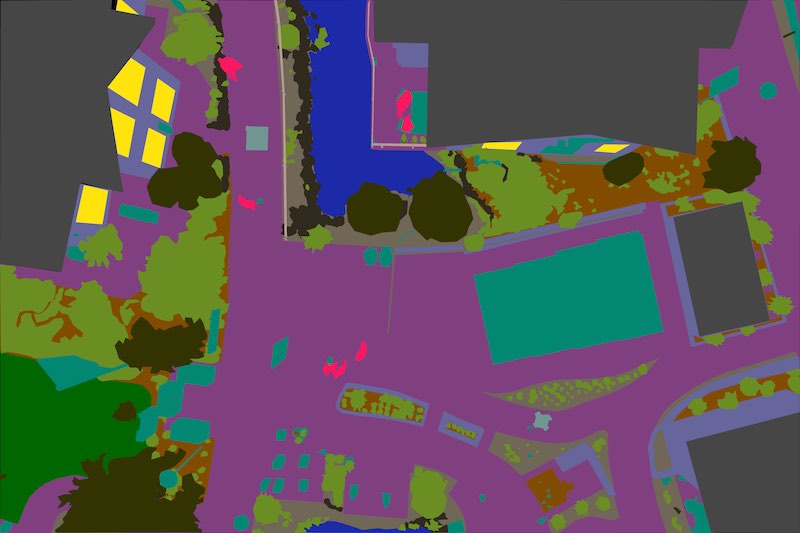}
\includegraphics[width=0.23\textwidth]{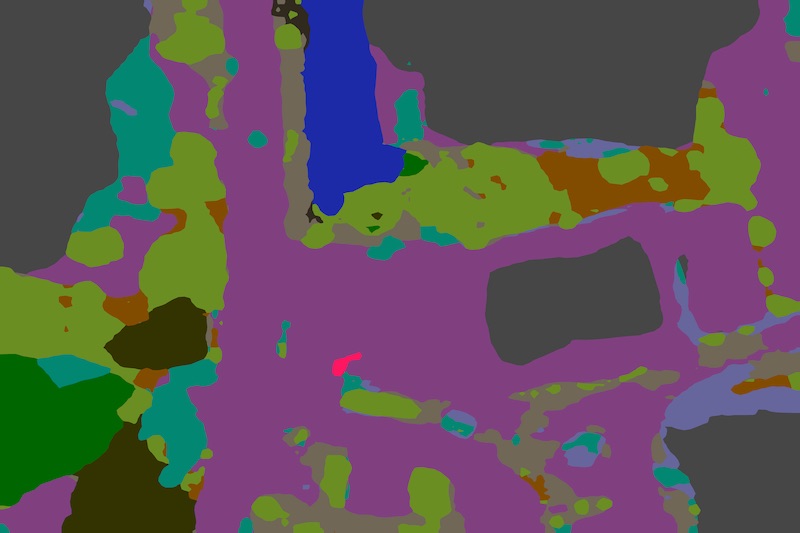}
\includegraphics[width=0.23\textwidth]{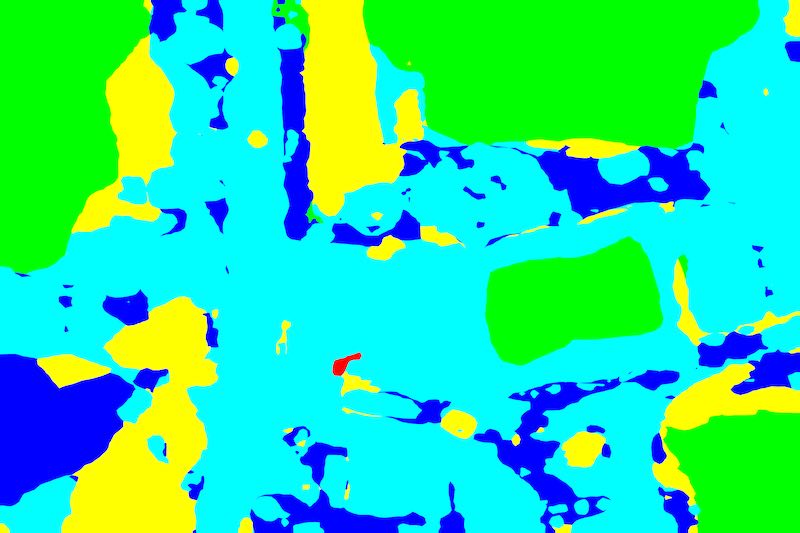}}
\end{subfigure}
\hfill
\hfill
\begin{subfigure}{\textwidth}
\centerline{d) \includegraphics[width=0.23\textwidth]{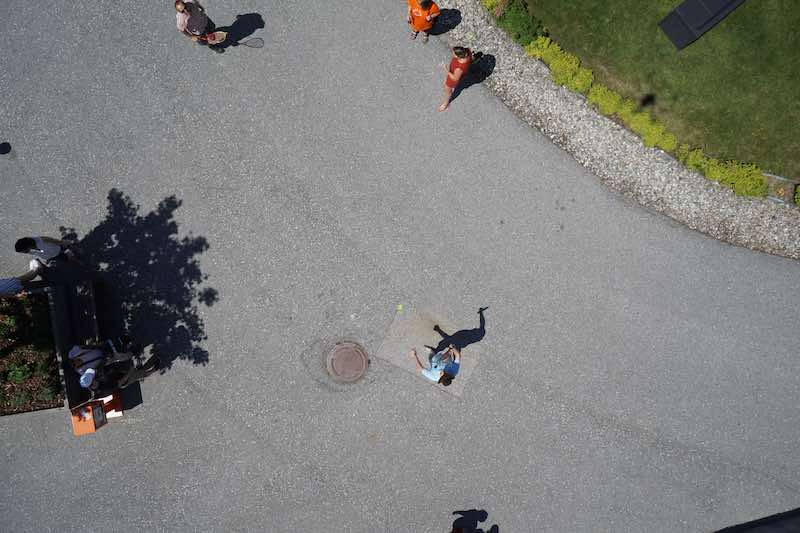}
\includegraphics[width=0.23\textwidth]{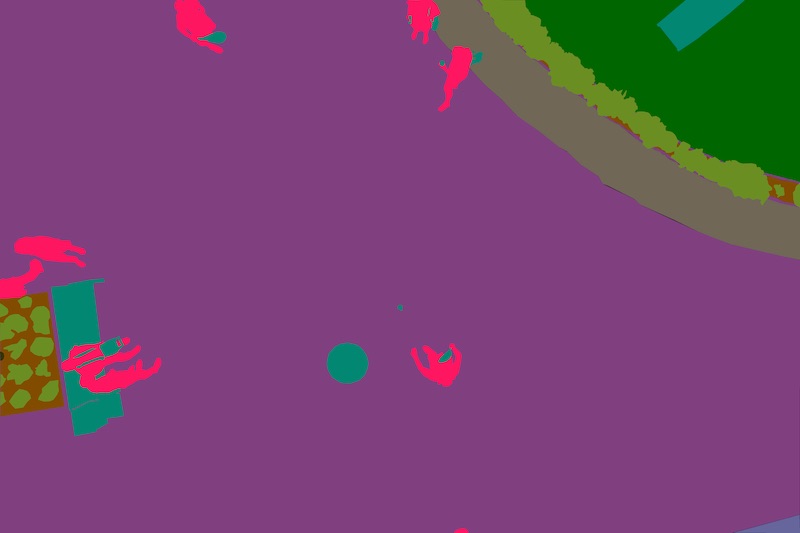}
\includegraphics[width=0.23\textwidth]{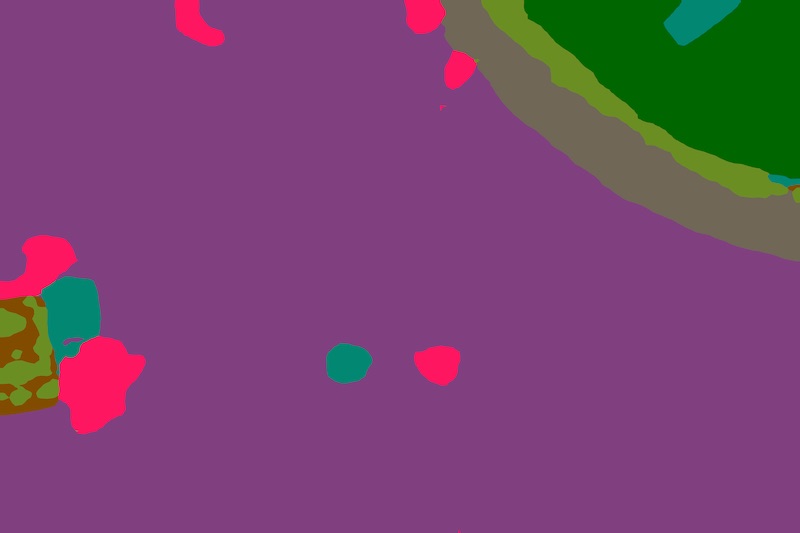}
\includegraphics[width=0.23\textwidth]{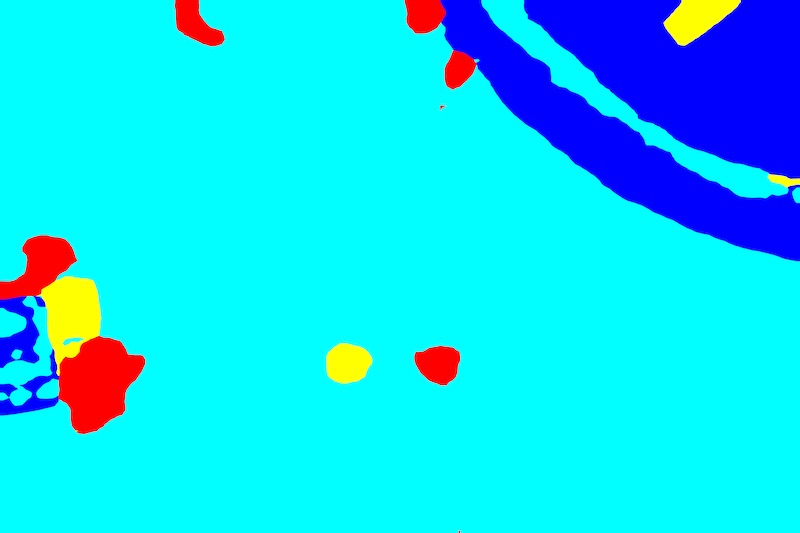}}
\end{subfigure}
\hfill
\hfill
\begin{subfigure}{\textwidth}
\centerline{e) \includegraphics[width=0.23\textwidth]{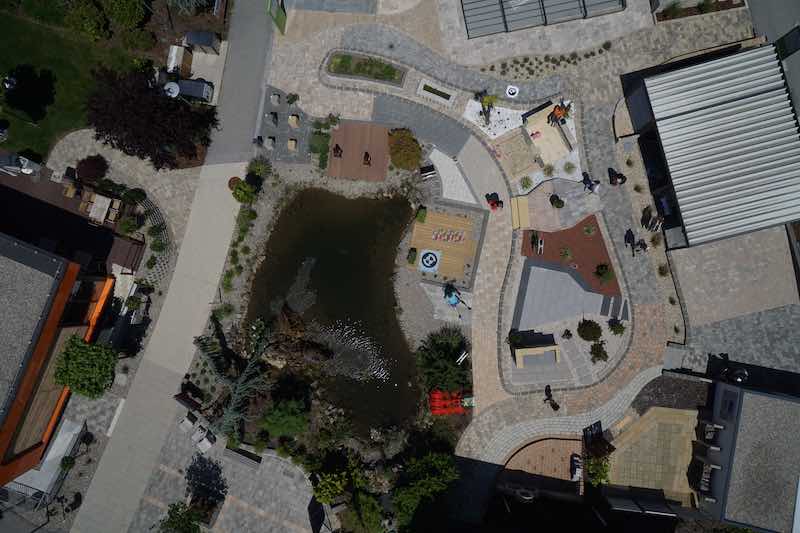}
\includegraphics[width=0.23\textwidth]{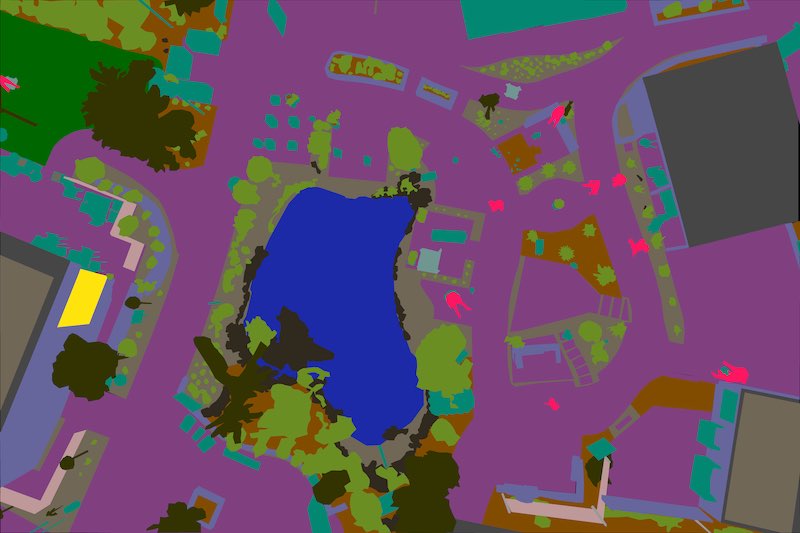}
\includegraphics[width=0.23\textwidth]{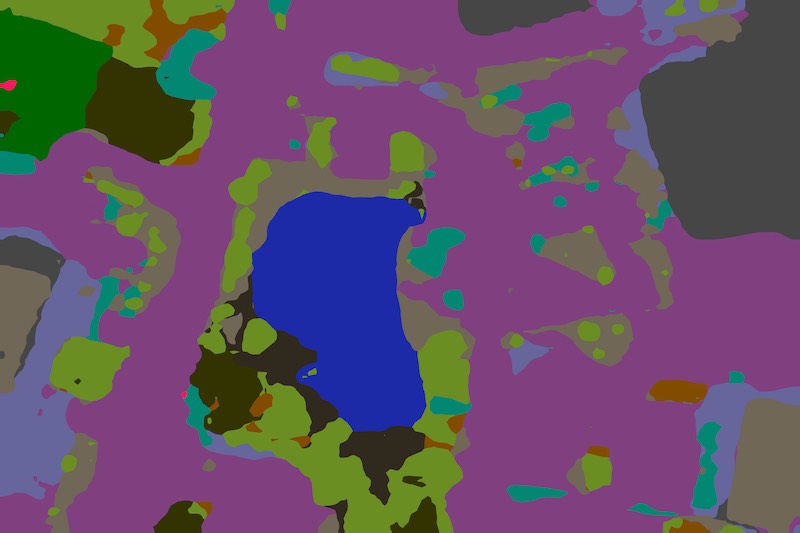}
\includegraphics[width=0.23\textwidth]{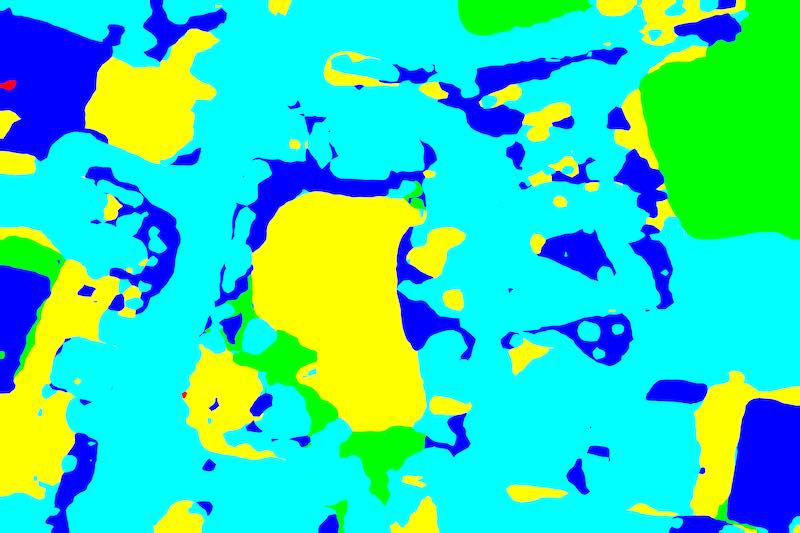}}
\end{subfigure}
\caption{Experimental results obtained for $5$ different %cases of study 
case studies within the test set. The first column presents the original image, the ground truth annotated image is depicted in the second column, and the third column contains the inferred semantic segmentation. The last column shows the risk levels, where the red color represents the highest risk, passing through orange, yellow, green, cyan, to blue, which represents the safest areas. Reference for colors in Table~\ref{tab:classes}.}
\label{fig:results}
\end{figure*}

%It should be noted there is a situation where the accuracy on the high risk levels drops dramatically. This situation is when aerial views are taken at high altitudes.
%Fig. \ref{fig:results} shows a variety of examples from the test dataset. It can be seen that from high altitude, categorization is not very detailed; classes are dispersed and mixed, especially for the high risk classes, as can be shown in the cases of study in the third and fifth rows.
Fig.~\ref{fig:results} shows a variety of examples from the test dataset, where an overall good performance is observed. One of the most notorious details from those is that aerial views from high altitude, high risk level classes are not very detailed, sometimes do not even appearing in the prediction, as can be shown in %the third and fifth examples of the cases of study where people is not detected properly.
rows $c)$ and $e)$.
Aside from requiring more training, this is caused by the small scale of these classes en the image and, as previously mentioned, their low count against low risk classes. This produces the considerable number of false negatives.
%It is worth noting that, as seen in the previously mentioned examples, from very high altitude the model did not correctly identify people in the scene.
This might seem problematic, but in a real scenario it would be expected that while approaching land, the model would start identifying people %(see the fourth row in Fig. \ref{fig:results})
as seen on $d)$. %Nevertheless, as previously mentioned, this could be produced because of the small size of the training set, and training with a larger dataset on higher-altitude images will also reduce this issue.
Again, this is a problem of generalization and could be addressed training the model with a larger dataset containing images taken from higher altitudes.
In contrast, images closer to the ground have fewer classes, and predictions are more accurate. This is important because in a landing attempt, the UAV will descend %and thus it will receive 
start receiving more confident information to adjust the last maneuvers. Future work can consider the altitude to produce a confidence index so that the UAV will take appropriate action based on it.%, i.e. take the final decision on where to land if UAV is at a certain altitude from a surface. 
In general, %we can appreciate 
this work show the huge potential of %semantic segmentation 
SS-based techniques to provide useful information %in this complex task.
to UAS to determine more confidently the best options to land in case of an emergency.

% \begin{figure*}[h!]
% \centering
% \includegraphics[width=\textwidth]{SSexamples.png}
% \caption{Experimental results obtained for $7$ different cases of study. The first column presents the original image, then, the ground truth annotated image is depicted in the second column, the third column contains the inferred semantic segmentation, while the last column showcases the risk levels, where red color represents maximum risk, passing through orange, yellow, green and cyan, until blue, which stand for the safest areas.}
% \label{fig:results}
% \end{figure*}

\begin{figure*}[h!]
\centering
\begin{subfigure}{\textwidth}
    \centerline{a) \includegraphics[width=0.23\textwidth]{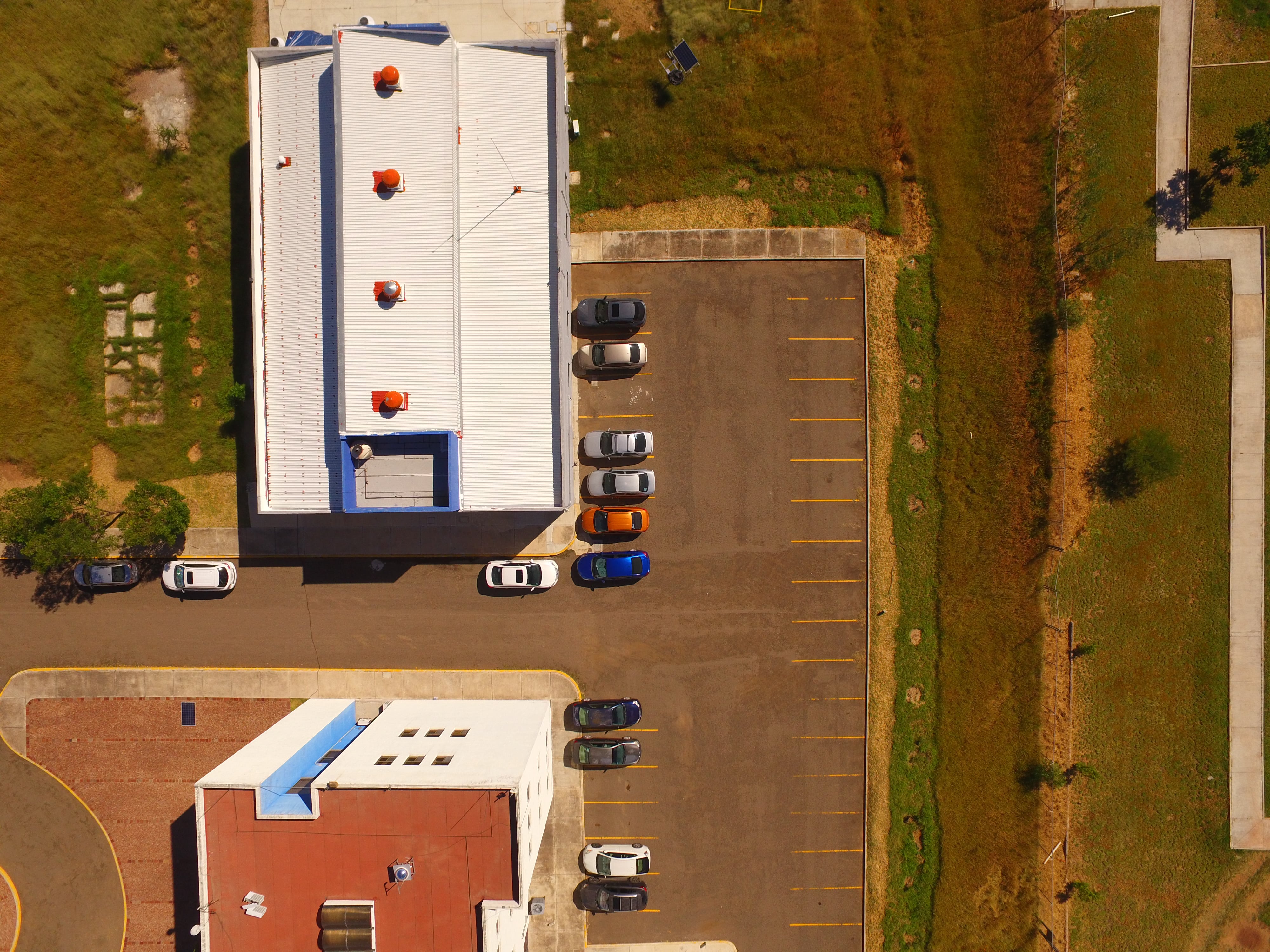}
    \includegraphics[width=0.23\textwidth]{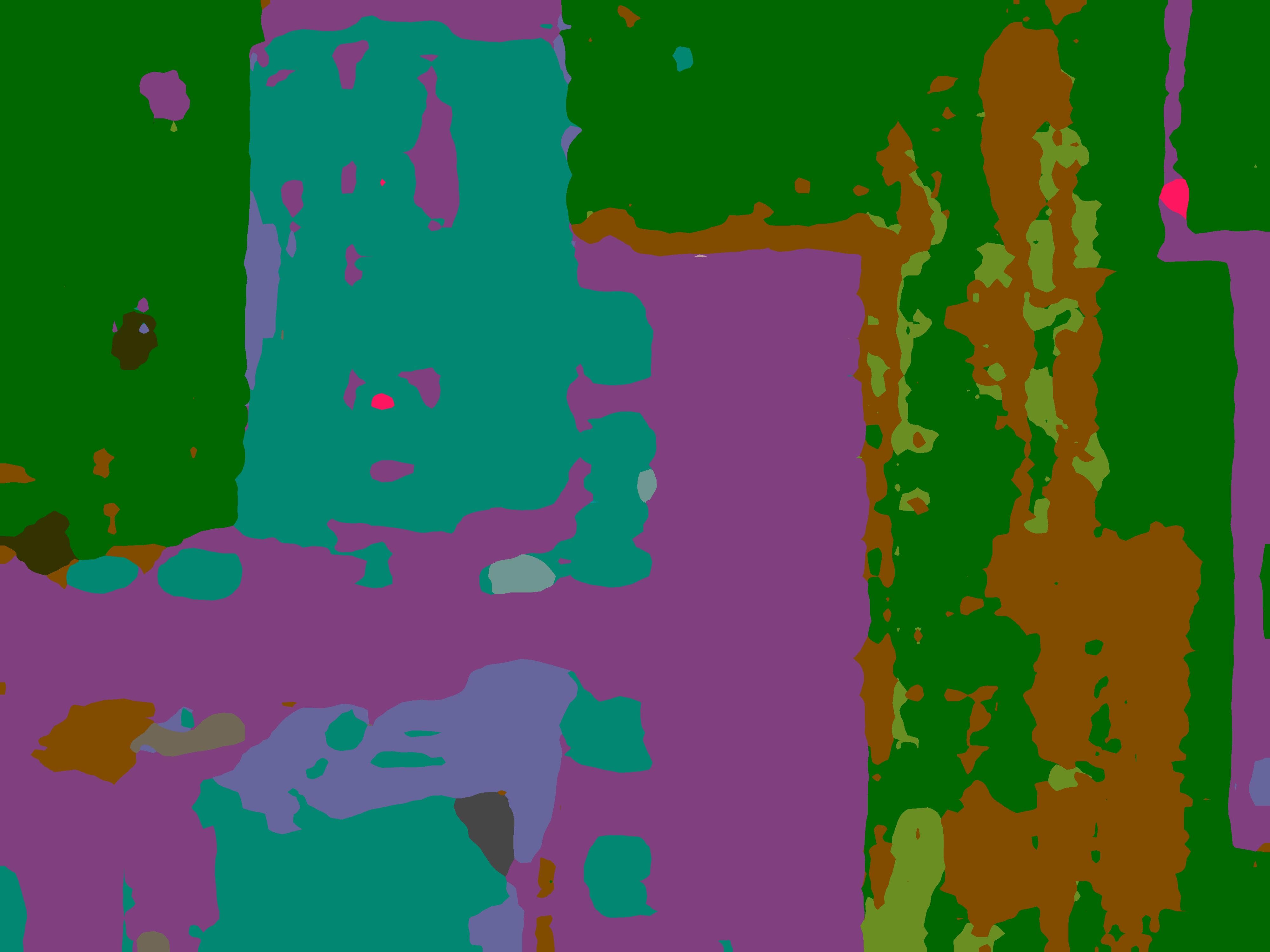}
    \includegraphics[width=0.23\textwidth]{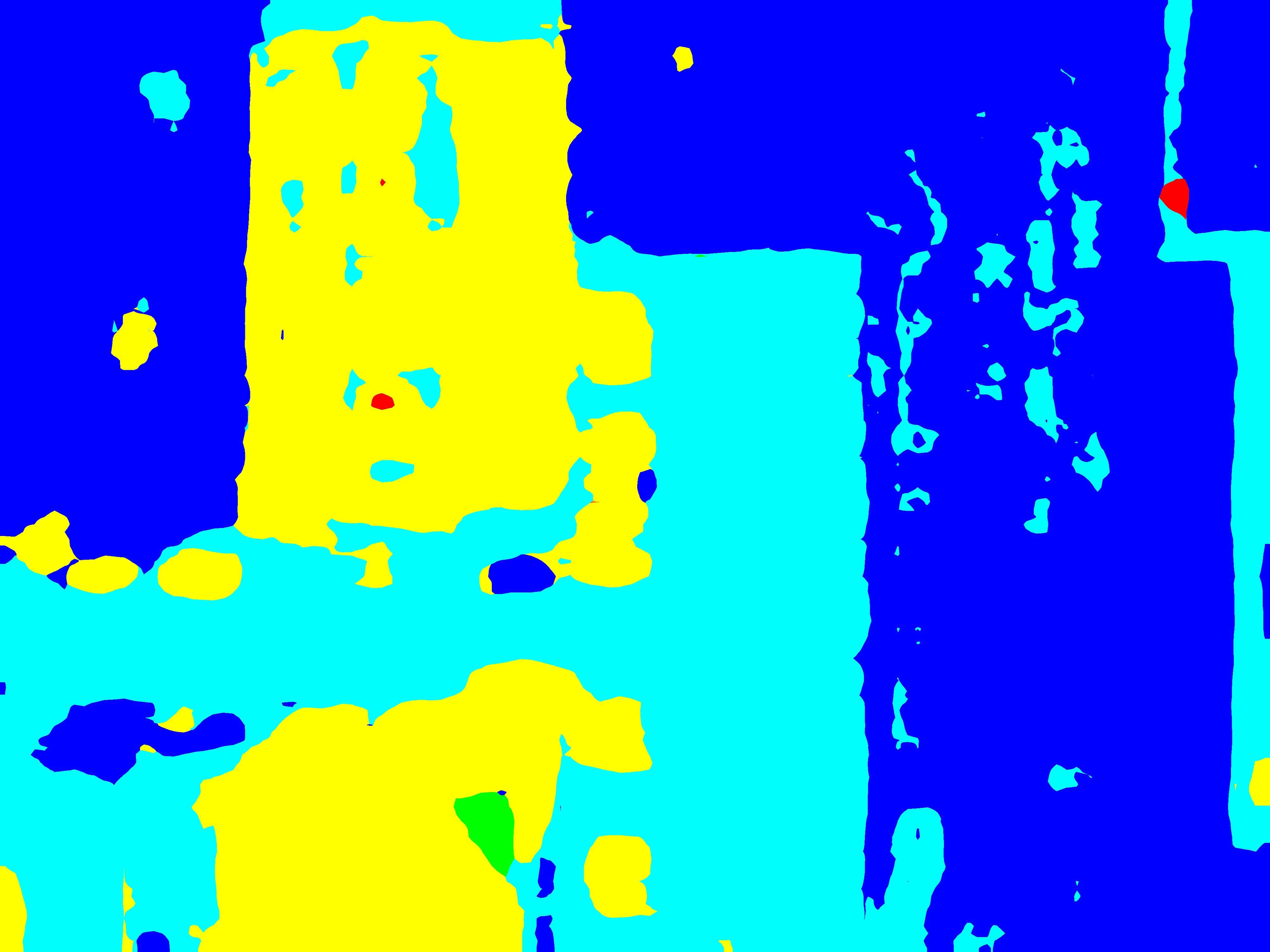}
    \includegraphics[width=0.23\textwidth]{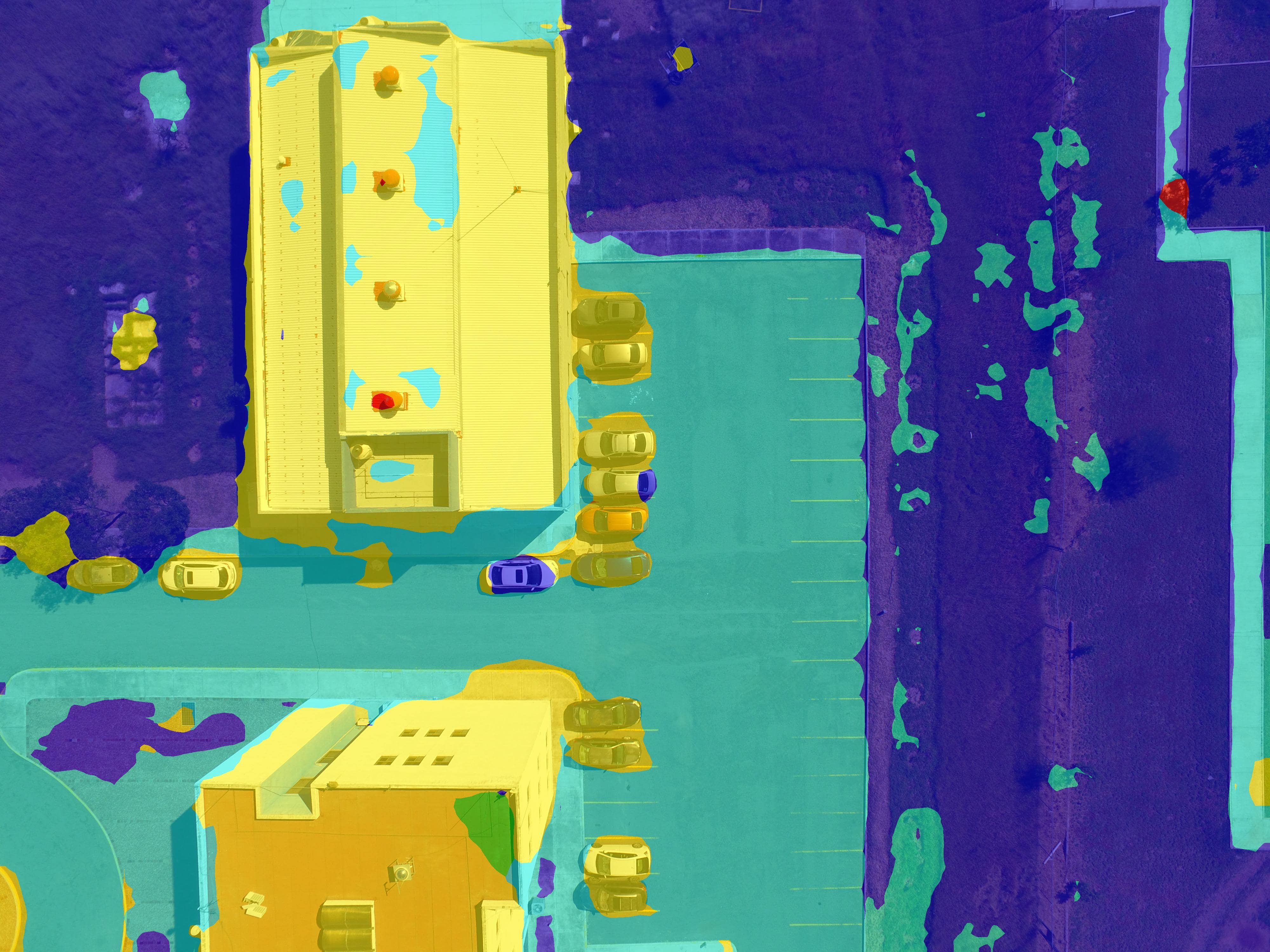}}
\end{subfigure}
\hfill
\hfill
\begin{subfigure}{\textwidth}
    \centerline{b) \includegraphics[width=0.23\textwidth]{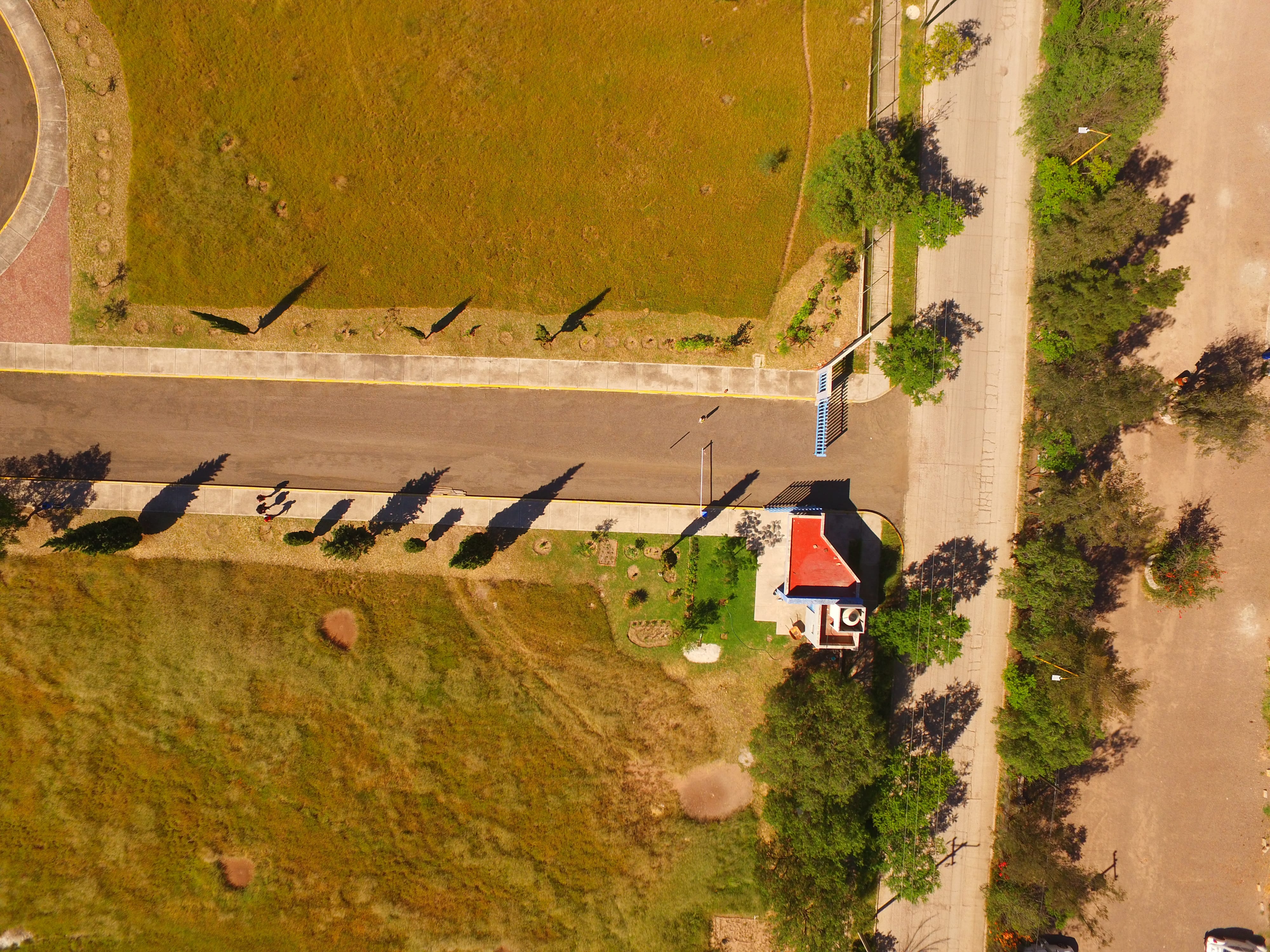}
    \includegraphics[width=0.23\textwidth]{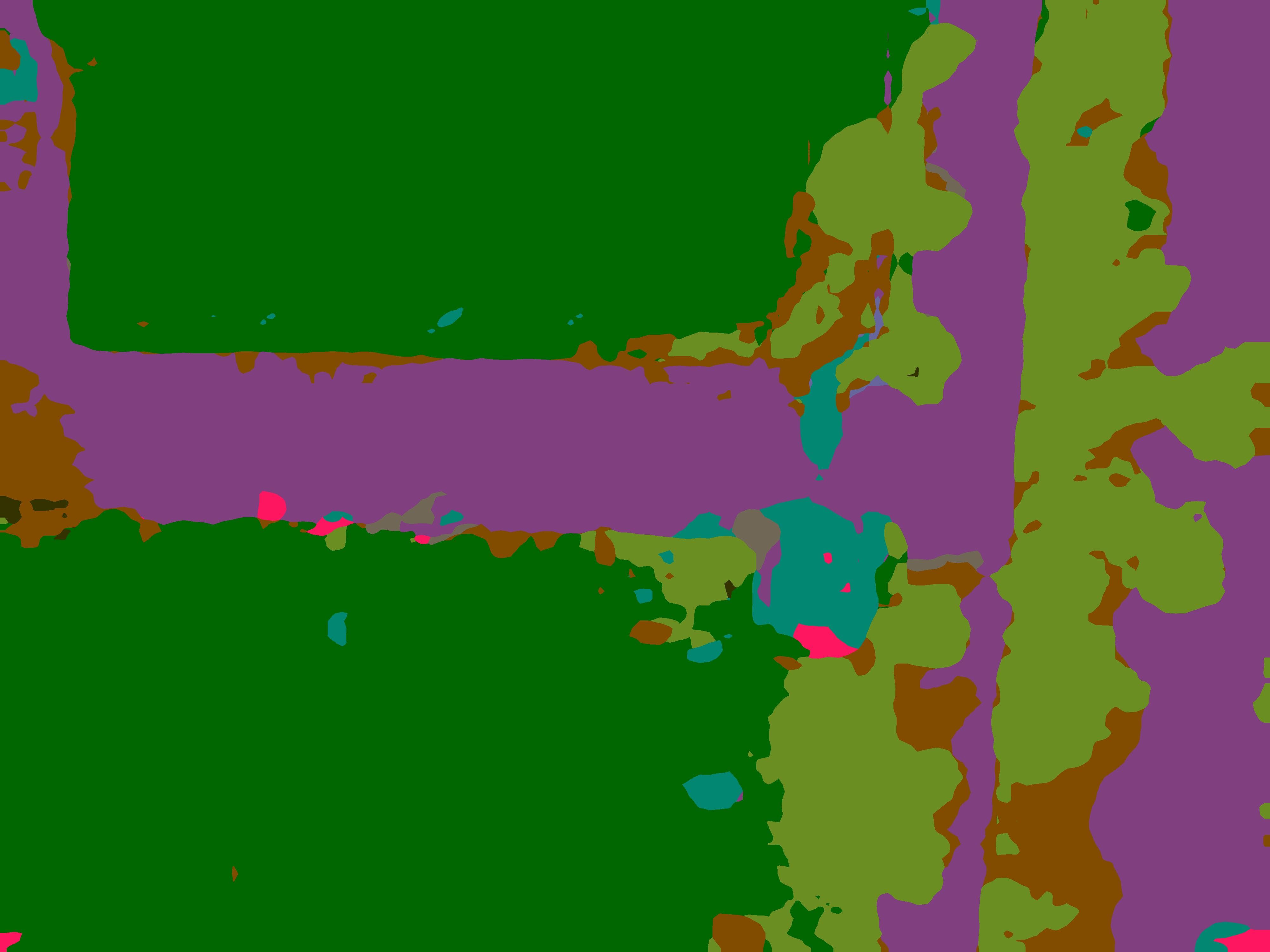}
    \includegraphics[width=0.23\textwidth]{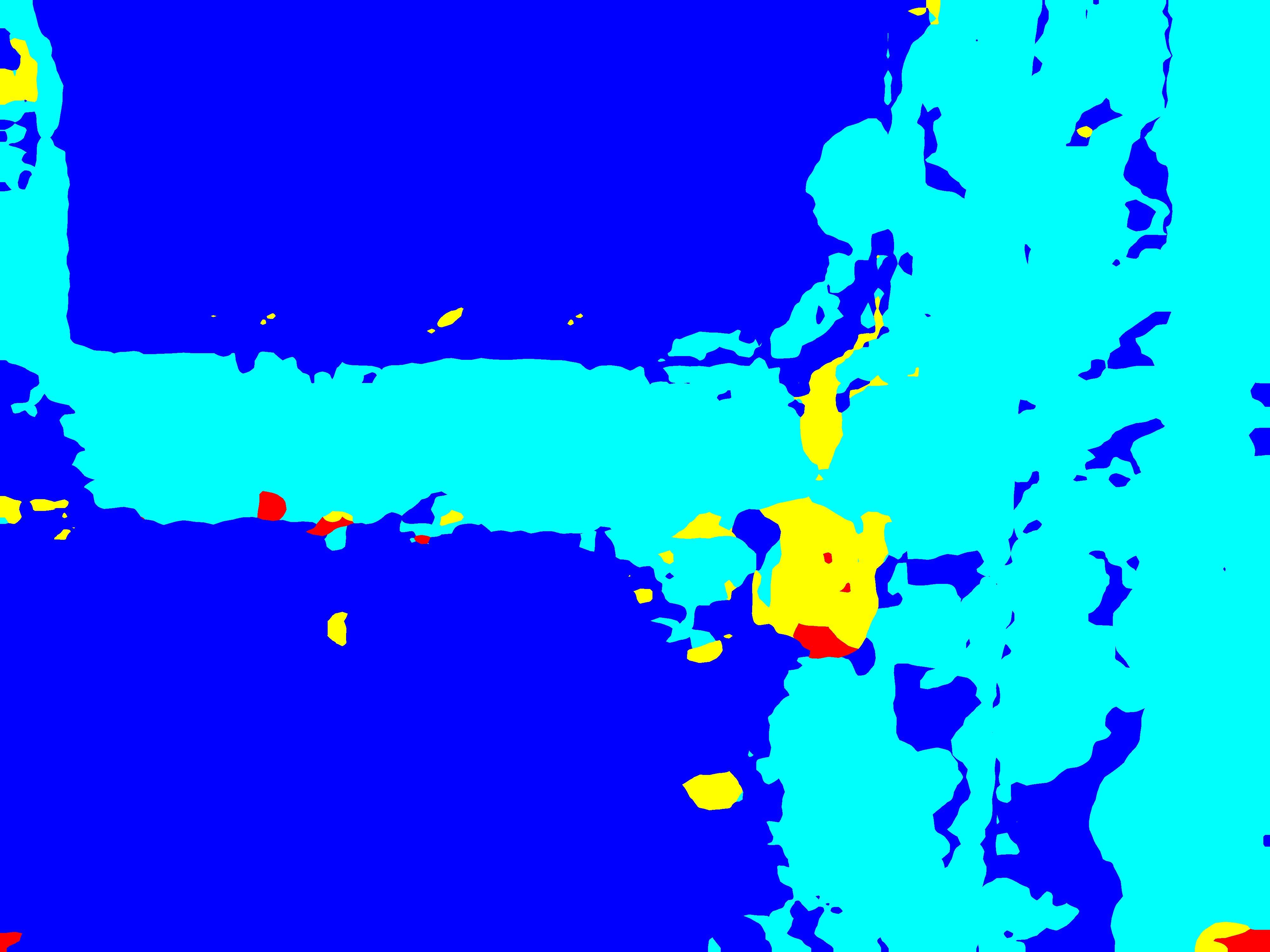}
    \includegraphics[width=0.23\textwidth]{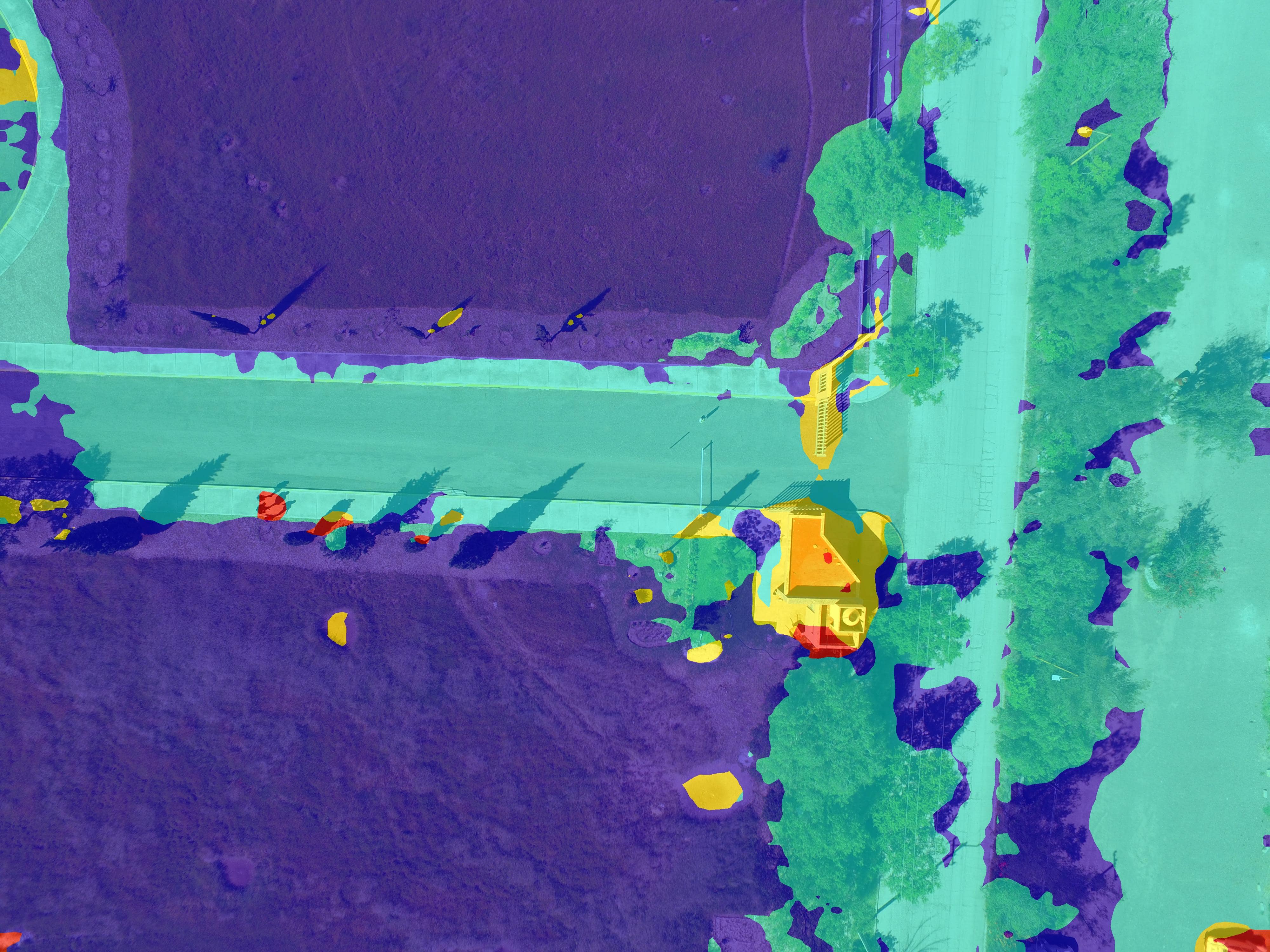}}
\end{subfigure}
\hfill
\hfill
\begin{subfigure}{\textwidth}
    \centerline{c) \includegraphics[width=0.23\textwidth]{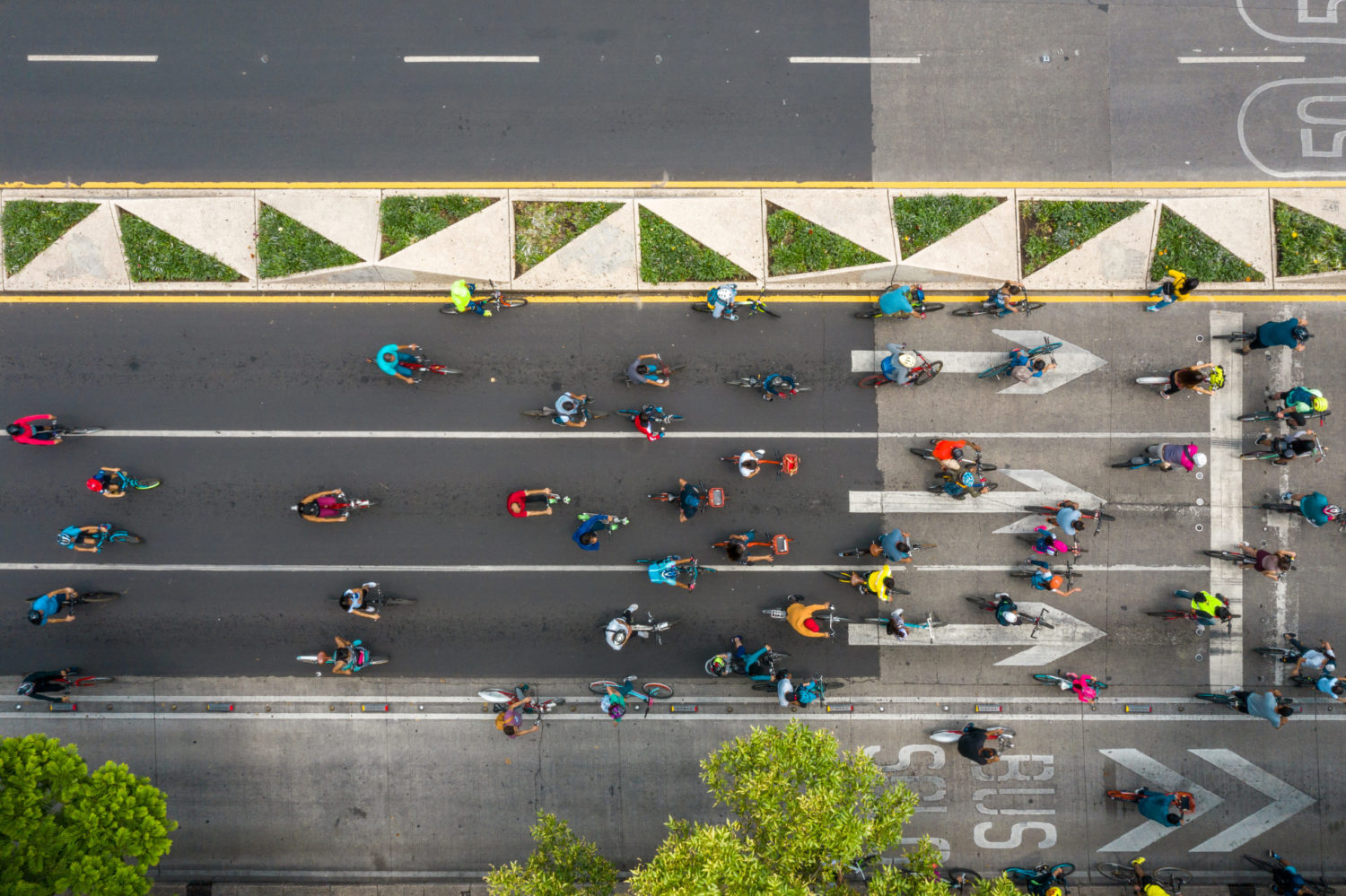}
    \includegraphics[width=0.23\textwidth]{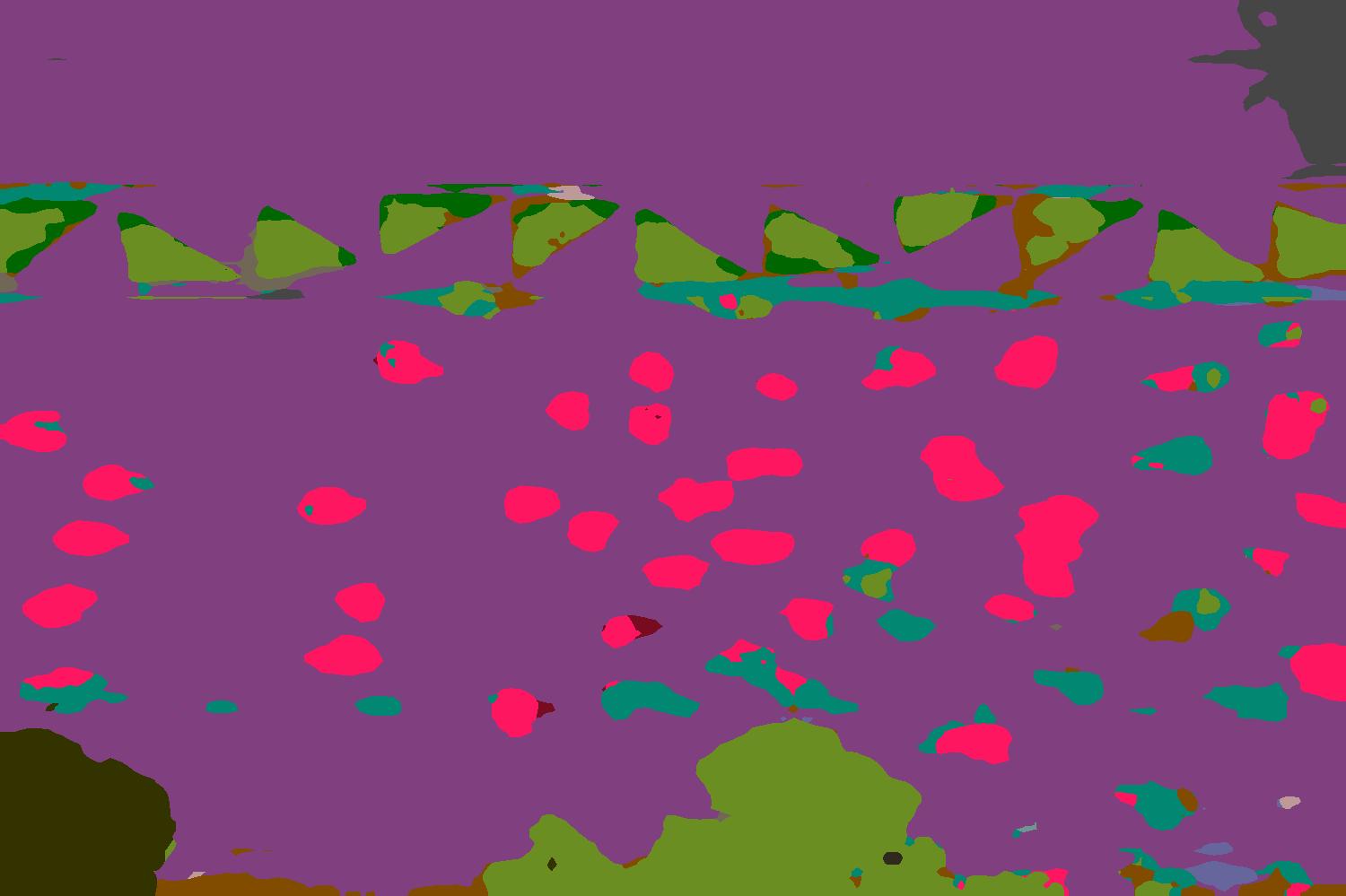}
    \includegraphics[width=0.23\textwidth]{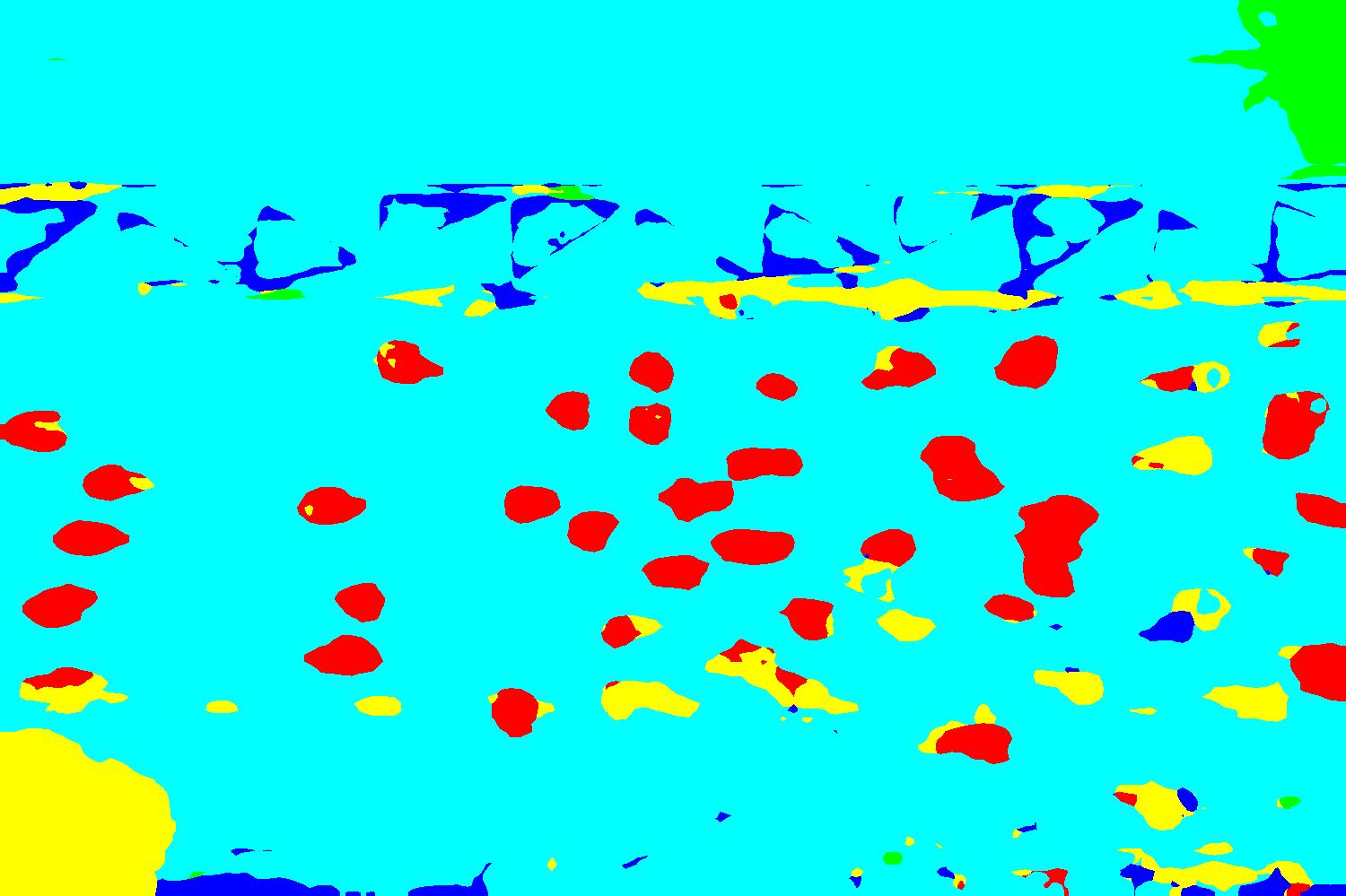}
    \includegraphics[width=0.23\textwidth]{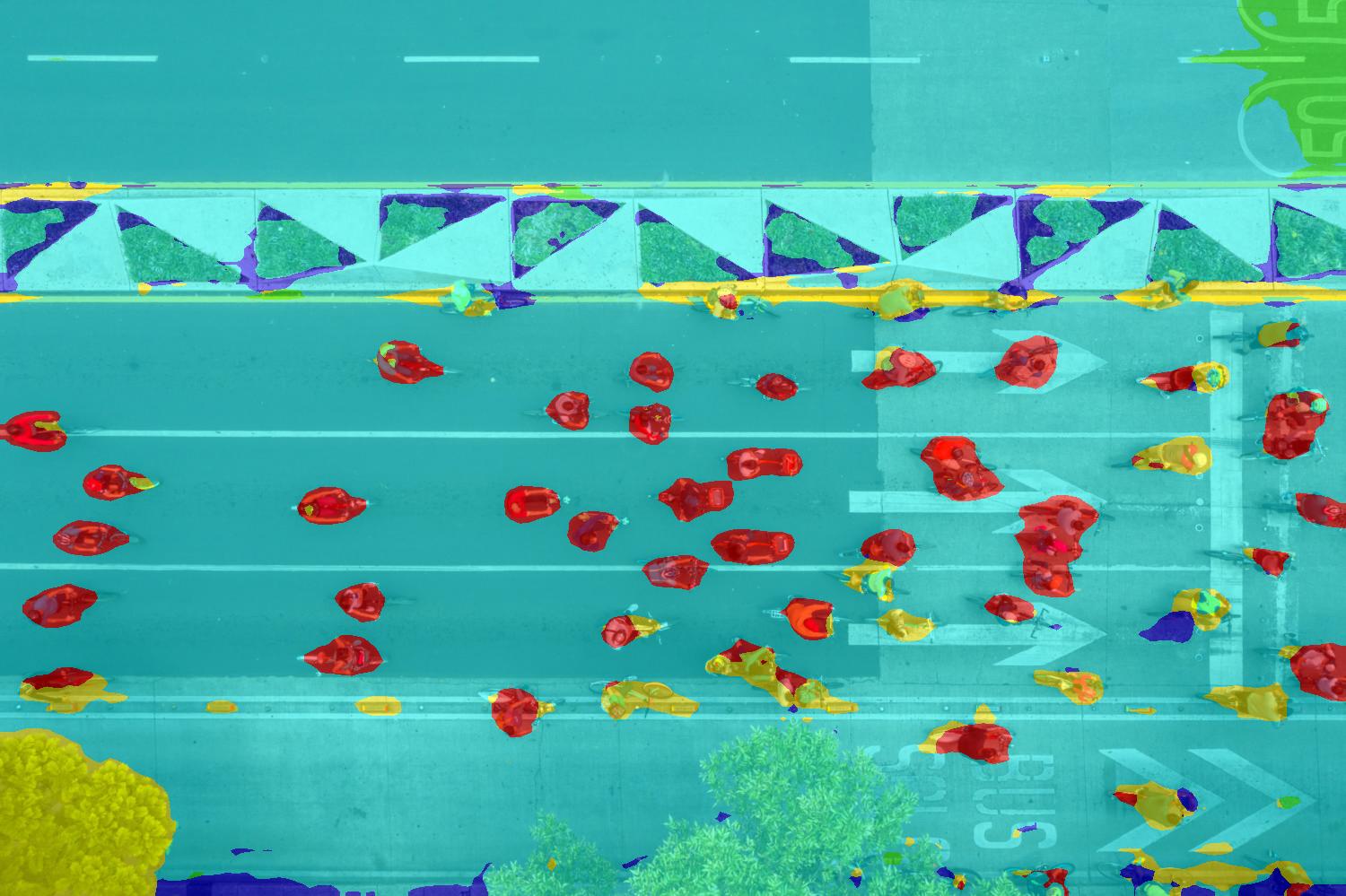}}
\end{subfigure}
\hfill
\hfill
\begin{subfigure}{\textwidth}
    \centerline{d) \includegraphics[width=0.23\textwidth]{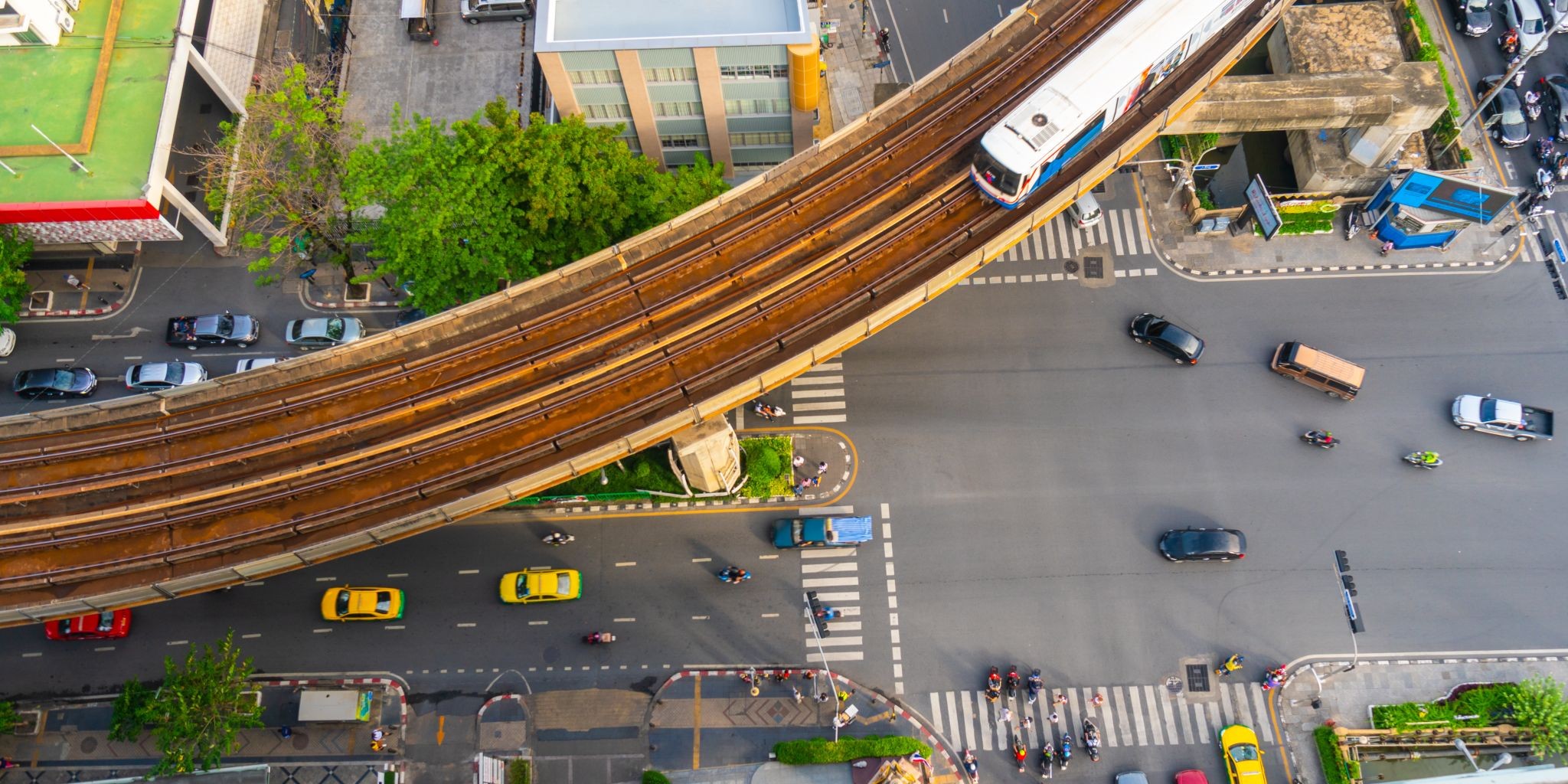}
    \includegraphics[width=0.23\textwidth]{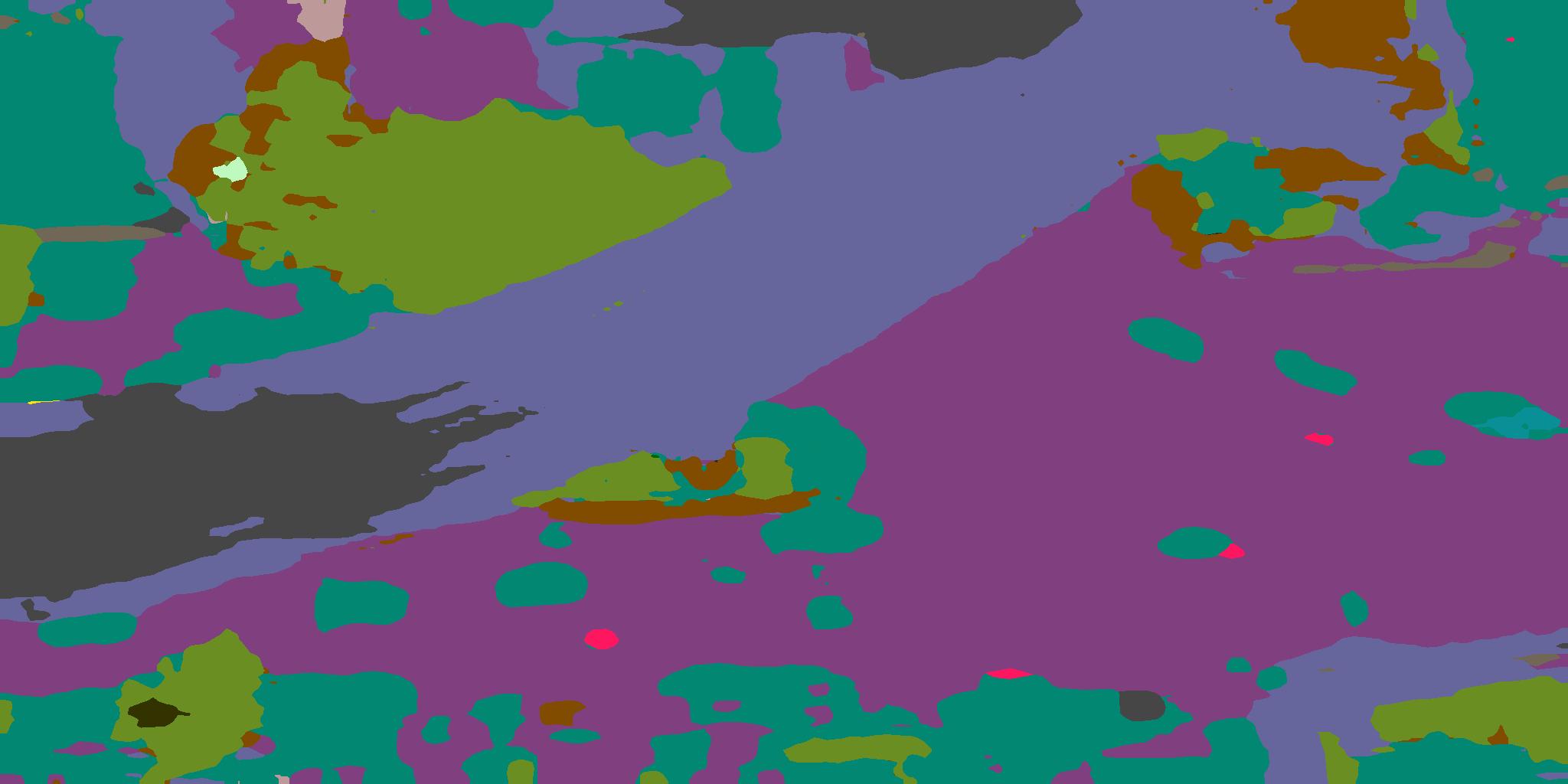}
    \includegraphics[width=0.23\textwidth]{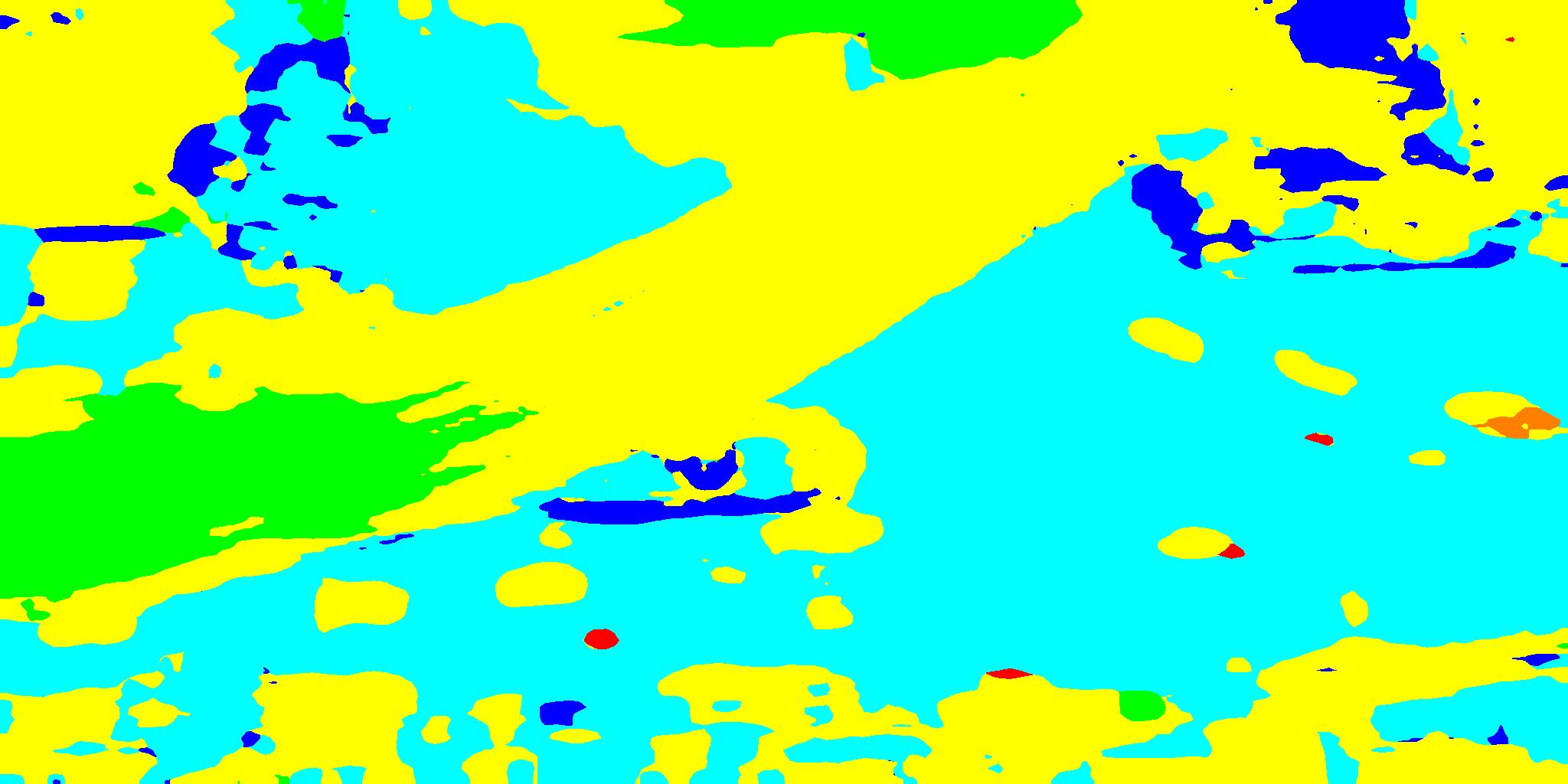}
    \includegraphics[width=0.23\textwidth]{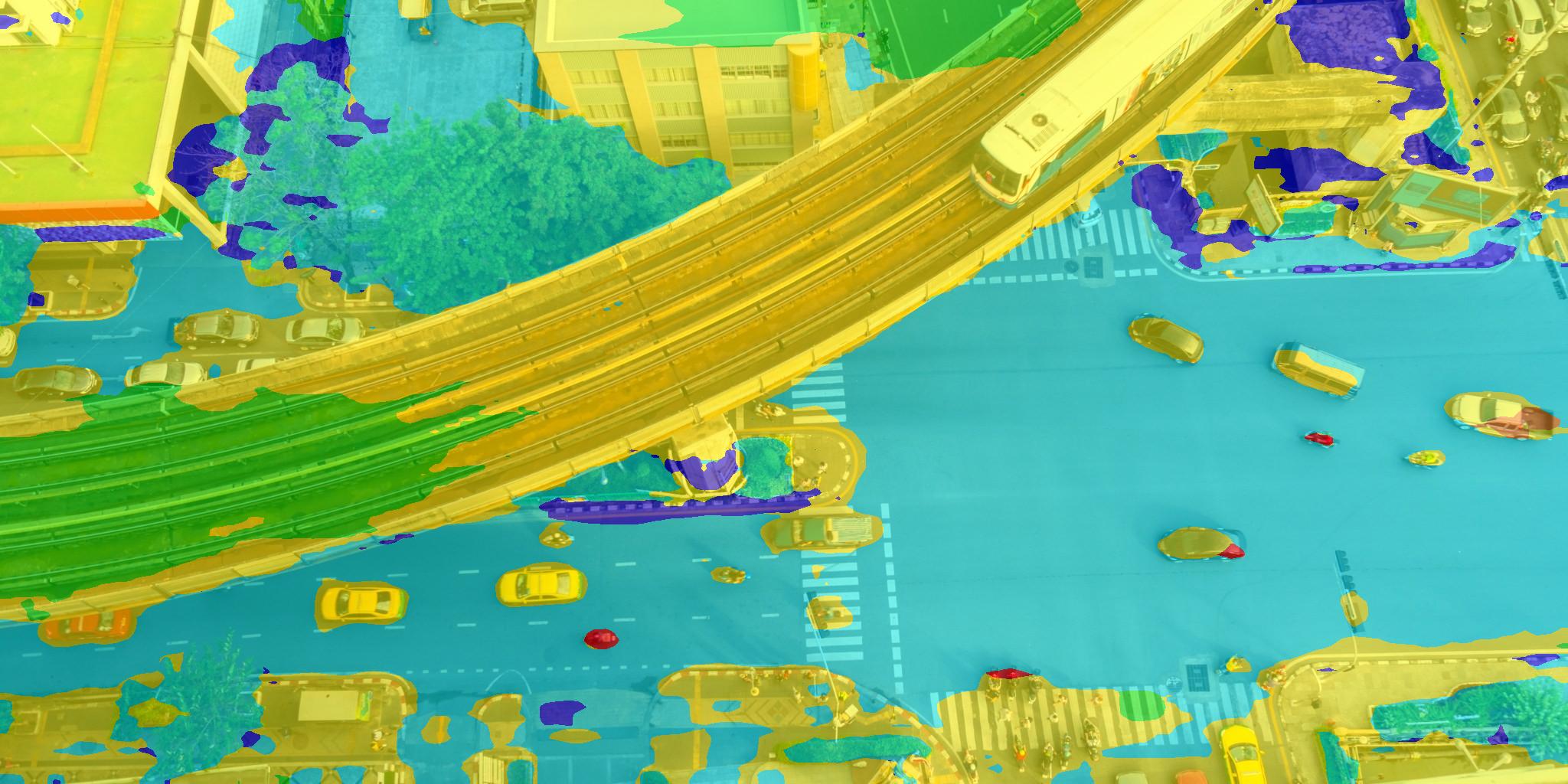}}
\end{subfigure}
%\hfill
%\hfill
%\begin{subfigure}{\textwidth}
%    \centerline{e) \includegraphics[width=0.23\textwidth]{VistaArea-81834529.jpg}
%    \includegraphics[width=0.23\textwidth]{VistaArea-81834529_SegMask.jpg}
%    \includegraphics[width=0.23\textwidth]{VistaArea-81834529_RiskMask.jpg}
%    \includegraphics[width=0.23\textwidth]{VistaArea-81834529_imgRisk.jpg}}
%\end{subfigure}
%\caption{Experimental results obtained for different cases of study using images with aerial views not included in the training set. The first column presents the original image, the inferred semantic segmentation map is depicted in the second column, and the third column contains the risk levels. The last column shows the original image masked with the risk level map to visualize directly the mapping done by the model.}
\caption{Experimental results obtained for different %study cases 
case studies using images with different aerial views not included in the training set. The first column presents the input image, then the inferred semantic segmentation map is depicted in the second column. Risk levels for the third column and a composition of the original image masked with the risk level map in the last column to visualize directly the mapping done by the model. Reference for colors in Table~\ref{tab:classes}.}
\label{fig:tests}
\end{figure*}

%To better analyze the results, and study the generalization capabilities of the trained model, 
Fig.~\ref{fig:tests} show some extra examples of inference from the trained model, but from aerial pictures in different scenarios not contained on the %training set
original dataset. In this case, the uncontrolled and different scenarios of the pictures produce a variety of results. First, it can be seen that for %the second example in row $e)$, 
test $c)$, the inferred results are quite good and very useful: cyclist are clearly find by the model and details of surroundings are well defined. Results could be this good because this picture is similar to the ones at the Semantic Drone Dataset. By contrast, other images show %some 
more important misclassifications that results in an incorrect risk level map. In the case at row $a)$, one car in the middle is identified as an AR-Marker, which puts it at very low risk. It also confuses an air exhaust as a person, although this does not cause any risk for the operation. These behaviors showcase the great complexity of the task of autonomous landing in real unstructured urban scenarios. The case in row %$c)$ 
$b)$ also signals some non existent people, and dirt at the right of the image is defined as paved-area. Finally, example %$e)$ 
$d)$ presents completely new classes not considered in the Semantic Drone Datasets, such as the train and its track. It can be seen that the model tries to categorize some features, as the train and its track as paved area and roof, which do not have the risk level that normally would be assigned to this objects. %Nevertheless, for the rest of the scene, the model presents a good prediction. Even with these errors, the model is able to provide with useful information which can be used for smat autonomous landing.
All of this only reinforces the need to train the model in a more extensive and diverse dataset, but even with errors, this proposal is able to provide useful information to preview the best areas for emergency AL.

Those extra examples demonstrate that the model and the categorization of the classes is a viable tool for UAVs when requiring to land autonomously in their surroundings. Some issues with the generalization could be mainly solved by training in a more diverse dataset. %Also, the discrete risk level map could be transformed into a map of probabilities of risk where big low risk areas could have a point where the risk is minimum for that particular area, and optimization strategies could be applied to find global minima in the whole scene, in which the UAV could quickly go to land in case of an emergency.
%Finally, it is also necessary to account for the dynamic of the environment: low risk areas can quickly turn into a hot-spot of pedestrians, so it will be necessary to track the areas through time to select the best overall Safe Landing Zone.
There are still some interesting approaches using this method, but for now this work presents a good alternative to assess the risk on the current image of a UAS which can be used to select the best available Safe Landing Zone.

\section{Conclusion}
\label{sec:conclusions}
In this article, a semantic segmentation-based risk assessment strategy for autonomous landing in complex unstructured urban scenarios was proposed. First, a state-of-the-art visual transformer network was chosen and trained for semantic segmentation image input from a UAS. Then, six different risk levels are obtained from the semantic segmentation %image 
model output, which %in turn can be used to smartly select the best available safe landing zones.
gives more certainty to find high risk areas to avoid in case of an autonomous landing.

%The proposed strategy was evaluated along different case studies, showcasing the huge potential of semantic segmentation-based strategies to provide important contextual information useful for decision making, particularly in emergency landing situations in complex urban environments. Furthermore, improving the resiliency of the system in case of system failure, for example, by using vision-based techniques, might help unleash the full potential of drones in urban civilian applications. All and all, the use of semantic segmentation-based techniques for autonomous landing in complex urban scenarios appears as a powerful tool, but a lot of effort is still required in order to obtain a reliable solution.
The method presented in this work was evaluated along different case studies, and showcases the huge potential of using semantic segmentation and risk grouping to provide important contextual information to the UAV about the ground risk in its surroundings, useful for decision making, particularly in emergency landing situations.% in complex urban environments.
The work presented here adds robustness to the system in case of an emergency landing is needed, improving %the 
its SORA score and %effectively paving the way ...
opening the way for applications in complex urban environments of being authorized.

Future work includes expanding the training dataset in order to capture more diversity of urban scenarios, camera angles, class variability and more height. Also, it is of interest to use the risk assessment information to choose a landing spot %in real-time, embedding the system on the processor onboard the UAV, and accounting for the dynamic nature of the scene.
using some formal decision process that accounts for uncertainty and other variables. Finally, it is planned to validate each module of the proposal in real settings, running the algorithms embedded in a UAV and finding a Safe Landing Zone.

\begin{credits}
\subsubsection{\ackname} %A bold run-in heading in small font size at the end of the paper is used for general acknowledgments, for example: This study was funded by X (grant number Y).
This work was supported by the Mexican National Council of Humanities, Science and Technology (Conahcyt), and the U.S. Office of Naval Research (ONR).

\subsubsection{\discintname}
% It is now necessary to declare any competing interests or to specifically state that the authors have no competing interests. Please place the statement with a bold run-in heading in small font size beneath the (optional) acknowledgments\footnote{If EquinOCS, our proceedings submission system, is used, then the disclaimer can be provided directly in the system.}, for example: The authors have no competing interests to declare that are relevant to the content of this article. Or: Author A has received research grants from Company W. Author B has received a speaker honorarium from Company X and owns stock in Company Y. Author C is a member of committee Z.
The authors have no competing interests to declare that are relevant to the content of this article.
\end{credits}

% ---- Bibliography ----
% BibTeX users should specify bibliography style 'splncs04'.
% References will then be sorted and formatted in the correct style.
%
\bibliographystyle{splncs04}
% \bibliography{mybibliography}

%\bibliographystyle{IEEEtran}
\bibliography{thereferences.bib}

%%%%%%%%%%%%%%%%%%%%%%%%%%
%\begin{thebibliography}{8}
%\bibitem{ref_article1}
%Author, F.: Article title. Journal \textbf{2}(5), 99--110 (2016)
%
%\bibitem{ref_lncs1}
%Author, F., Author, S.: Title of a proceedings paper. In: Editor, F., Editor, S. (eds.) CONFERENCE 2016, LNCS, vol. 9999, pp. 1--13. Springer, Heidelberg (2016). \doi{10.10007/1234567890}
%
%\bibitem{ref_book1}
%Author, F., Author, S., Author, T.: Book title. 2nd edn. Publisher, Location (1999)
%
%\bibitem{ref_proc1}
%Author, A.-B.: Contribution title. In: 9th International Proceedings on Proceedings, pp. 1--2. Publisher, Location (2010)
%
%\bibitem{ref_url1}
%LNCS Homepage, \url{http://www.springer.com/lncs}, last accessed 2023/10/25
%\end{thebibliography}

\end{document}